\documentclass{article}
\usepackage{amssymb}

\usepackage[preprint]{corl_2025} 

\usepackage{amsmath, amsfonts}
\usepackage{graphicx} 
\usepackage{xcolor}

\usepackage{algorithm}
\usepackage{algorithmic}

\usepackage{multirow}
\usepackage{booktabs}
\usepackage{subcaption}

\usepackage{wrapfig}
\usepackage{subcaption}
\usepackage{placeins}
\usepackage{comment}

\usepackage{amsthm}

\newcommand\rahel[1]{{\color{black} #1}}

\newcommand{\cross}[1][1pt]{\ooalign{%
  \rule[1ex]{1ex}{#1}\cr
  \hss\rule{#1}{.7em}\hss\cr}}

\DeclareMathOperator*{\argmin}{arg\,min}

\title{ZipMPC: Compressed Context-Dependent MPC Cost via Imitation Learning}

\author{
  Rahel Rickenbach$^{*,1}$ \\
  \texttt{rrahel@ethz.ch} \\
  \And
  Alan A. Lahoud$^{*,2}$ \\
  \texttt{alan.lahoud@oru.se} \\
  \And
  Erik Schaffernicht$^{2}$\\
  \texttt{erik.schaffernicht@oru.se} \\
  \And
  Melanie N. Zeilinger$^{\cross,1}$\\
  \texttt{mzeilinger@ethz.ch} \\
  \And
  Johannes A. Stork$^{\cross,2}$\\
  \texttt{johannesandreas.stork@oru.se} \\
}

\begin{document}
\maketitle

\renewcommand{\thefootnote}{}

\footnotetext{\hspace{-1.8ex}\textsuperscript{1}Institute for Dynamic Systems and Control, ETH Zurich, Switzerland. \textsuperscript{2}Center for Applied Autonomous Sensor Systems (AASS), Örebro University, Sweden. \textsuperscript{*}These authors contributed equally (shared first authorship). \textsuperscript{\cross}These authors contributed equally (shared last authorship).}


\vspace{-2.0em}
\begin{abstract}

The computational burden of model predictive control (MPC) limits its application on real-time systems, such as robots, and often requires the use of short prediction horizons.
This not only affects the control performance, but also increases the difficulty of designing MPC cost functions that reflect the desired long-term objective.
This paper proposes ZipMPC, a method that imitates a long-horizon MPC behaviour by learning a compressed and context-dependent cost function for a short-horizon MPC. It improves performance over alternative methods, such as approximate explicit MPC and automatic cost parameter tuning, in particular in terms of i) optimizing the long-term objective; ii) maintaining computational costs comparable to a short-horizon MPC; iii) ensuring constraint satisfaction; and iv) generalizing control behaviour to environments not observed during training.
For this purpose, ZipMPC leverages the concept of differentiable MPC with neural networks to propagate gradients of the imitation loss through the MPC optimization. 
We validate our proposed method in simulation and real-world experiments on autonomous racing. ZipMPC consistently completes laps faster than selected baselines, achieving lap times close to the long-horizon MPC baseline. In challenging scenarios where the short-horizon MPC baseline fails to complete a lap, ZipMPC is able to do so. In particular, these performance gains are also observed on tracks unseen during training.\\ Videos and code are available at \href{https://zipmpc.github.io/}{zipmpc.github.io/}.

\end{abstract}



\section{Introduction}
Model predictive controllers (MPCs) rely on optimization-based control actions that minimize an MPC cost while respecting environmental and system-specific constraints. This allows for their successful deployment in a range of safety-critical control applications, such as robot locomotion \citep{grandia2019, romualdi2022, sleiman2021unified, villarreal2020mpc} and manipulation \cite{incremona2017mpc,arcari2023bayesian}, autonomous racing \citep{kabzan2019, costa2023, liniger2015optimization}, or biomedical robotics \cite{rodriguez2017anfis,dunkelberger2023hybrid,sun2023knee} and applications \citep{hovorka2004, rivadeneira2020impulsive}. However, \rahel{to make MPC computationally tractable, its prediction horizon is commonly chosen significantly smaller than the horizon required for task completion, resulting in an information gap. Therefore,} MPC applications face two challenges; (i) the selection of a tractable MPC prediction horizon length, i.e., the time frame considered during optimization, (ii) 
the design of an MPC cost function that accurately reflects the desired performance objective. 
\rahel{Applying MPC with a long prediction horizon, good performance can commonly be achieved with an intuitively and manually designed MPC cost, as the control policy is optimized considering vast knowledge of future environmental features, such as upcoming curvature in autonomous racing applications or glucose intake in insulinic pumps.} 
However, given that the computation time increases with the number of optimization variables, i.e., the horizon length \cite{sawma2018effect,dogruer2024optimizing}, applications with long-horizon MPCs are often not real-time feasible. Therefore, MPC with a short prediction horizon length is generally more common. 
\rahel{Short-horizon MPC formulations, on the other hand, render the design of the cost function challenging and often suffer from reduced performance due to the limited incorporation of future information.}
\rahel{Available approaches to address these challenges include approximate explicit MPC (eMPC), which learns efficient closed-form policies to imitate MPC behaviour, learning the cost-to-go, which learns a cost term to represent the information gap until task completion, as well as automatic cost parameter tuning via inverse optimal control (IOC) or Bayesian optimization (BO). While each of them offers individual advantages, from constraint satisfaction and generalization capabilities to computational efficiency, the goal of this paper is to propose a framework that combines these strengths.}  

\begin{figure}[h]
    \centering
    \includegraphics[width=\linewidth]{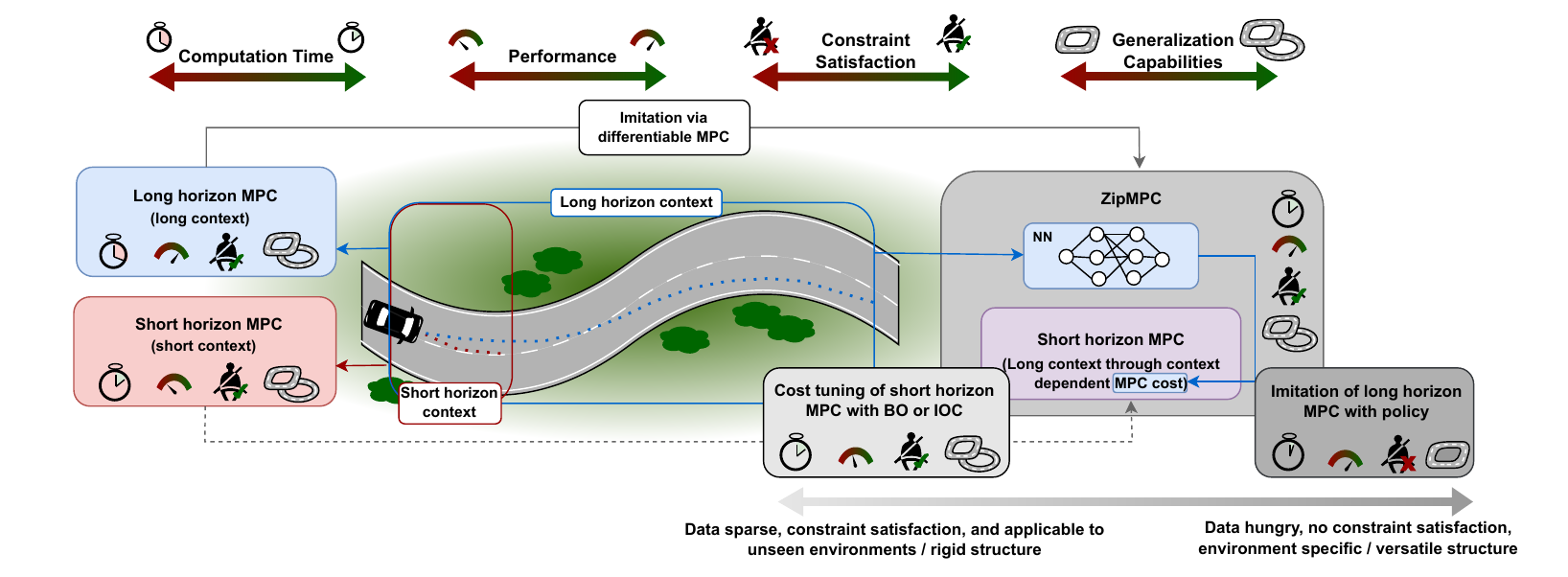}
    \vspace{-1.5em}
    \caption{ZipMPC imitates long-horizon MPC behaviour by learning a compressed and context-dependent cost \rahel{for} a short-horizon MPC. It thereby combines the faster inference times of eMPC with the generalization and constraint recognition capabilities \rahel{of} BO and IOC.}
    \label{fig:contribution}
\end{figure}

\paragraph{Contribution.} We propose ZipMPC, an algorithm designed to combine the computational advantages of short MPC horizons with the richer contextual information associated with longer horizons. Specifically, as depicted in Figure \ref{fig:contribution} for racing applications, ZipMPC employs a learnable function, e.g., using neural networks (NN), to encode long-horizon contextual information, such as track curvature, into a shorter horizon MPC cost. Leveraging differentiable MPC techniques, ZipMPC enables offline imitation learning of optimal long-horizon MPC solutions by propagating informative gradients through the MPC optimization procedure. By utilizing the computational efficiency of NN inference, our method places itself between eMPC and autonomous parameter tuning of MPC cost parameters by combining their strengths, i.e., it reduces the computational time, \rahel{while considering the dynamics to ensure constraint satisfaction. 
Due to the fact that we learn a cost function instead of a behavioural cloning policy, ZipMPC demonstrates a generalization capability, leading to improved performances in environments both seen and unseen during training.} 
We validate ZipMPC in simulation and on real-world (hardware) autonomous racing scenarios.
 
\vspace{-1.5em}
\vspace{-1.5em}

\paragraph{Related Work.}
Automatic tuning of MPC cost parameters is generally explored using BO to minimize the unknown mapping of MPC cost parameter values to a desired performance measurement \cite{Fröhlich2022}, employing optimality conditions to estimate the unknown objective vector from available optimal demonstrations via IOC \cite{FengShunLin:2021,abazar2020}, \rahel{or employing RL techniques \cite{gros2019data,zanon2019practical}.}
\rahel{While increasing the performance when applied to long-horizon MPC, they do not alleviate the computational demand. When applied to short-horizon MPC, current methods are not designed to include additional information to overcome the information gap and fall short in achieving a satisfying imitation behaviour. An exception describes contextual BO \cite{frohlich2022contextual}, which, however, does not scale well with context dimension. }
\rahel{To reduce computational complexity}, several works have explored \rahel{to reduce the horizon and to learn the missing information until task completion, commonly referred to as learning the cost-to-go 
\cite{abdufattokhov2021learning, seel2022convex, orrico2024building,rosolia2019costtogo}, as well as} replacing MPCs with trained NNs (eMPC) due to their quick inference. For instance, \cite{lucia2018deep, alsmeier2025imitation, kumar2018deep} use deep learning models to approximate MPC solutions through offline training, aiming to replicate MPC control policies in real-time. However, these approaches struggle with generalization \rahel{to unseen environments during training}. Furthermore, as NNs naturally fail to output structured predictions \cite{bakir2007predicting}, e.g., predictions that satisfy MPC hard constraints, \rahel{eMPC commonly does not offer constraint satisfaction properties}.
Recently, there has been growing interest in combining NNs with constrained optimization \cite{kotary2021end}. A key direction uses mathematical solvers as layers within deep learning models to enforce hard constraints in the outputs. These methods often involve differentiating solver outputs with respect to their inputs or parameters \cite{amos2017optnet, agrawal2019differentiable}. 
Specifically, methods for the integration of MPCs  
as a differentiable block within deep learning frameworks have been developed by \cite{drgona2020differentiable}, which propose adding ReLU-based penalty terms in the loss function to penalize constraint violations, or \cite{amos2018differentiable}, which use iterative Linear Quadratic Regulator (iLQR) as an approximation to MPC, and differentiates through the iLQR's fixed point by leveraging gradients through KKT conditions. \rahel{Differentiable MPCs are, for example, currently employed for the imitation of human behaviour \cite{acerbo2023evaluation}. With our proposed framework, we leverage them for reducing computational demands.}  


\section{MPC Preliminaries}

In this section, we introduce the MPC formulation serving as the presented framework’s basis. 
Let $x(k+1) = f(x(k),u(k),\zeta(k))$ represent some context-dependent discrete-time dynamics at time step $k$, where $x \in \mathbb{R}^n$, $u \in \mathbb{R}^m$ and $\zeta \in \mathbb{R}^{a}$ denote the state, control-input and environmental context variable, respectively. With $N$ indicating the prediction horizon, a general MPC controller \rahel{solves} the following problem at time step $k$ given the current state $x(k)$:
\begin{equation}
\label{eq:mpclh}
\begin{aligned}
 [X_{k,N}^*, U_{k,N}^*] := \mathit{MPC}_{N}(\rahel{C_N}, x(k)) = \argmin_{\{x_i\}_{0}^{N+1}, \{u_i\}_{0}^{N}} \sum_{i=0}^{N} \ell(x_i, u_i,\rahel{c_i})  \\
    \text{s.t.} \quad x_{0} = x(k), \quad x_{i+1} = f(x_i,u_i,\zeta_{i}), \quad x_i \in \mathcal{X}, \quad u_i \in \mathcal{U},  \quad i=0,\hdots,N.
\end{aligned}
\end{equation}
The function $\ell$ defines the stage cost 
\rahel{with parameters $c_i \in \mathbb{R}^{b}$, summarized in $C_{N} := \{c_i\}_{0}^{N}$.}
Furthermore, state and control-input constraints are indicated with $\mathcal{X} \subseteq \mathbb{R}^n$ and $\mathcal{U}\subseteq \mathbb{R}^m$, respectively. Solving Eq. \ref{eq:mpclh} returns an optimal state and control-input trajectory for all prediction steps of horizon $N$, denoted as $X_{k,N}^*:= \{x_i^*\}_{0}^{N+1}$ and $U_{k,N}^*:= \{u_i^*\}_{0}^{N}$, where $x_i^* \in \mathcal{X}$ and $u_i^* \in \mathcal{U}$ indicate the optimal state and input at prediction step $i$. Only the first optimal control-input $u_0^*$ is applied before the optimization is reinitialized and resolved, referred to as receding horizon control. 
To address theoretical and practical issues potentially resulting from the finite horizon implementation, such as missing \textit{convergence} and \textit{recursive feasibility} guarantees \citep{kouvaritakis2016model,grune2017nonlinear}, an additional cost term and constraint set on $x_{N+1}$, known as terminal ingredients, are often used to imitate infinite horizon behaviour. However, given that the design of terminal ingredients can be nontrivial, theoretical investigations on how to overcome their necessity for long-horizon MPCs have found interest in the recent literature \cite{boccia2014,soloperto2023outputtracking,Köhler2022outputregulation}. This further simplifies \rahel{the MPC cost design with long-horizon length and motivates the use of an MPC formulation without terminal ingredients in this work. 
Note, however, that the inclusion of a terminal cost would be straight forward.}

\section{ZipMPC: Compressed MPC Cost via Imitation Learning}

\rahel{In this section, we first define the underlying problem formulation, and then introduce ZipMPC, our method to learn an MPC cost via imitation learning by efficiently compressing long-horizon context information into a short-horizon MPC cost, followed by its detailed algorithm.}

\subsection{Problem Formulation}
\rahel{Motivated by the improved performance of an intuitively tuned MPC with increasing horizon length (see Appendix \ref{apx:laptime})}, we formulate the following problem.
Let $N_L$ and $N_S<N_L$, both $\in \mathbb{Z}$, represent a long enough (but expensive) and a short (but computationally efficient) prediction horizon length, respectively. We denote $[X_{k,N_L}^*, U_{k,N_L}^*] := \mathit{MPC}_{N_L}(\rahel{C^{M}_{N_L}}, x(k))$ as the optimal state and control-input variables given a long-horizon $N_L$ with well tuned cost parameters \rahel{$C^{M}_{N_L}$} and \rahel{current state $x(k)$}. Similarly, we denote $[X_{k,N_S}^{\theta}, U_{k,N_S}^{\theta}] := \mathit{MPC}_{N_S}(\rahel{C^{\theta}_{N_S}}, x(k))$ as the optimal state and control-input variables given a short-horizon $N_S$ with learned cost parameters $\rahel{C^{\theta}_{N_S}}$, where $\theta$ represents learnable parameters (e.g., NN weights).
We aim to find $\theta^*$ that minimizes an imitation loss \rahel{$\mathcal{L}$} between the predicted reference trajectories of the long-horizon MPC, and the predicted trajectories of the short-horizon MPC with the inferred costs, i.e., we aim to imitate the behaviour of a well-tuned long-horizon MPC. The problem formulation is mathematically defined as
\begin{equation}
\label{eq:problem}
\begin{aligned}
 \theta^* := \argmin_{\theta} \mathbb{E}_{k} [\mathcal{L}([X_{k,N_S}^{\theta}, U_{k,N_S}^{\theta}], [X_{k,N_L}^*, U_{k,N_L}^*])],
\end{aligned}
\end{equation}
where $k$ encodes information of initial states sampled from a set of feasible states, further detailed in the next section.

\subsection{Learning framework}

\begin{figure}[h]
    \vspace{-1.3em}
    \centering
    \includegraphics[width=0.8\linewidth]{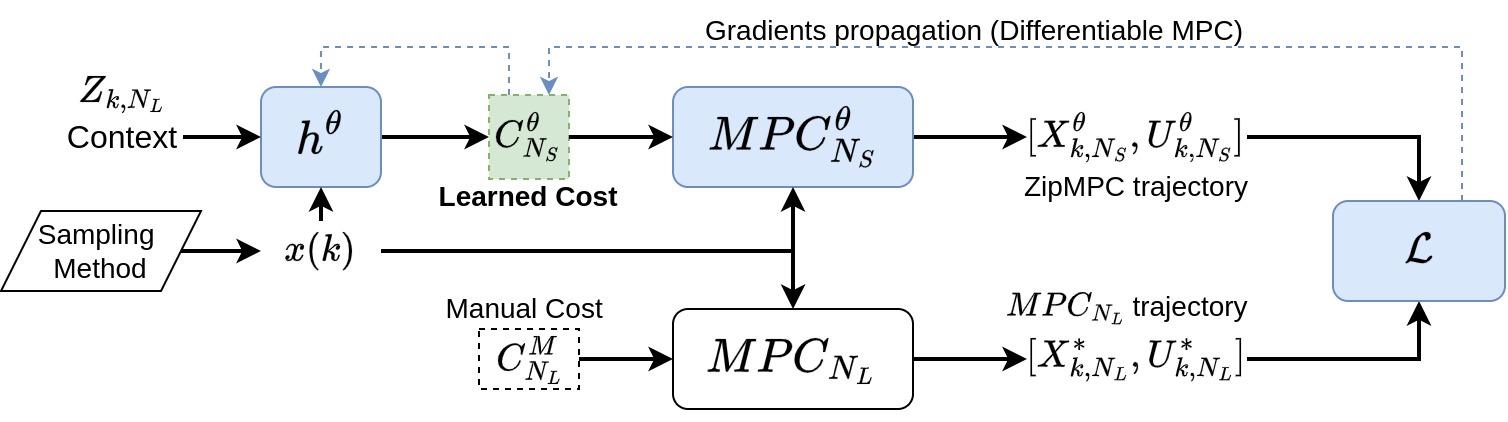}
    \caption{Simplified diagram of our learning approach. A NN parametrized by $\theta$ \rahel{receives a sample state $x(k)$ and corresponding context $Z_{k,N_{L}}$} as an input to predict cost parameters of \rahel{$\mathit{MPC}_{N_S}^{\theta}$. The weights $\theta$ are then optimized to minimize an imitation loss computed between the first $N_S$ steps of the trajectory yielded by $\mathit{MPC}_{N_L}$ and the inferred trajectory of $\mathit{MPC}_{N_S}^{\theta}$.}}
    \label{fig:framework}
    \vspace{-1.0em}
\end{figure}

To overcome the information gap between the optimization procedures of $\mathit{MPC}_{N_S}$ and $\mathit{MPC}_{N_L}$, 
we propose modeling the short-horizon MPC cost parameters $\rahel{C^{\theta}_{N_S}}$ as functions of the current state $x(k)$ and \rahel{some} contextual information \rahel{sequence} $Z_{k,N_L}$ that \rahel{ideally} contains \rahel{at least} as much information as provided to $\mathit{MPC}_{N_L}$.  
Thus, we define a parametric nonlinear function \rahel{$h^{\theta}$}, learned using NN weights $\theta$, such that \rahel{$C^{\theta}_{N_S} = h^{\theta}(x(k), Z_{k,N_L})$}. The trajectory $[X_{k,N_S}^{\theta}, U_{k,N_S}^{\theta}]$ is then provided by the optimization procedure in $\mathit{MPC}_{N_S}^{\rahel{\theta}} := \mathit{MPC}_{N_S}(\rahel{C^{\theta}_{N_S}}, x(k))$, and the loss of Eq. \ref{eq:problem} is computed against the reference trajectory $[X_{k,N_L}^*, U_{k,N_L}^*]$. Since trajectories from $\mathit{MPC}_{N_L}$ and $\mathit{MPC}_{N_S}^{\theta}$ differ in horizon length, the imitation loss is computed using only the first \rahel{$N_S$} steps of both trajectories. This ensures that $\mathit{MPC}_{N_S}^{\theta}$ anticipates behaviour within these steps, which would be unaccounted for without explicitly incorporating the long-horizon context $ Z_{k,N_L}$ into our learning framework. A sampling method provides initial states $x(k)$ for both the \rahel{parametric function $h^{\theta}$} and the MPCs, \rahel{$\mathit{MPC}_{N_S}^{\theta}$ and $\mathit{MPC}_{N_L}$}. Although various sampling methods could be used, \rahel{a simple strategy is to apply} uniform random sampling from a ``\rahel{feasible}" region.
\rahel{We ensure that every considered $x(k)$ in the learning process is ``feasible" by only keeping the ones that adhere to $\mathit{MPC}_{N_L}$ feasibility.} A high-level diagram of our learning approach is illustrated in Figure \ref{fig:framework}.

In order to solve the optimization in Eq. \ref{eq:problem} using gradient-based algorithms, we need to propagate gradients from the imitation loss, i.e., $\mathcal{L}([X_{k,N_S}^{\theta}, U_{k,N_S}^{\theta}], [X_{k,N_L}^*, U_{k,N_L}^*])]$, to the NN weights $\theta$, which is formulated by the following equation:
\begin{equation}
\label{eq:chain_rule}
\frac{\partial \mathcal{L}}{\partial \theta} = \frac{\partial \mathcal{L}}{\partial [X_{k,N_S}^{\theta}, U_{k,N_S}^{\theta}]} \cdot \frac{\partial [X_{k,N_S}^{\theta}, U_{k,N_S}^{\theta}]}{\partial \rahel{C^{\theta}_{N_S}}} \cdot \frac{\partial \rahel{C^{\theta}_{N_S}}}{\partial \theta}.
\end{equation}
It is well known that the bottleneck of the gradient backpropagation lies in the second term of the right hand side of the above equation \cite{donti2017task, wilder2019melding, lahoud2024learning}, since it requires to propagate the gradients through the optimization procedure, which in our case is the $\mathit{MPC}_{N_S}^{\theta}$ optimization. Therefore, we ensure the flow of informative gradients of the MPC solution with respect to the MPC cost parameters by leveraging differentiable MPC methods. \rahel{In particular}, we use the differentiable method of \cite{amos2018differentiable} as a tool, as it fits to general MPC cost formulations by extending the KKT-based differentiation \cite{amos2017optnet} to handle non-convexities by iteratively solving a convex quadratic approximation using iterative Linear Quadratic Regulator.

\paragraph{Detailed Algorithm.}

\begin{wrapfigure}{r}{0.56\textwidth} 
    \vspace{-1.3em} 
    \begin{minipage}{\linewidth}
        \begin{algorithm}[H]
            \caption{ZipMPC}
            \textbf{Inputs:} \\
            Initial NN weights $\theta$ of a NN function $h$.\\
            Context information $Z$.\\
            Initial MPC manual cost parameters $c^{M}$.\\ 
            Horizon lengths $N_S$ and $N_L$.\\
            \vspace{-0.8em}
            \begin{algorithmic}[1]
            \label{alg:ml_mpc}
                \STATE $C^{M}_{N_S} \gets \text{EXPAND}(c^{M}, N_S)$ %
                \STATE $C^{M}_{N_L} \gets \text{EXPAND}(c^{M}, N_L)$ %
                \FOR{$i = 1$ to $N_{ITERATIONS}$}
                    \STATE $x(k) \gets \text{SAMPLE}()$ 
                    \STATE $\Delta C^{\theta}_{N_S} \gets h^{\theta}(x(k),Z_{k, N_L})$ 
                    \STATE $C_{N_S}^{\theta} \gets C^{M}_{N_S} + \Delta C_{N_S}^{\theta}$ 
                    \STATE $[X^{\theta}_{k, N_S}, U^{\theta}_{k, N_S}] \gets \mathit{MPC}_{N_S}^{\theta}(C_{N_S}^{\theta}, x(k))$ 
                
                    \STATE $[X^*_{k,N_L}, U^*_{k,N_L}] \gets \mathit{MPC}_{N_L}(C^{M}_{N_L}, x(k))$ 
                    
                    \STATE $\mathcal{L} \gets \mathrm{MSE}_{N_D}([X^*_{k, N_L}, U^*_{k, N_L}], [X^{\theta}_{k, N_S}, U^{\theta}_{k, N_S}])$
                    \STATE $\theta \gets \text{Adam}(\theta, \mathcal{L})$ 
                \ENDFOR
            \end{algorithmic}
        \end{algorithm}
    \end{minipage}
    \vspace{-4ex} 
\end{wrapfigure}

The step-by-step process of ZipMPC within a gradient-based learning process is presented in Algorithm \ref{alg:ml_mpc}. It follows the diagram in Figure \ref{fig:framework} with some additional details that we highlight as follows: i) We assume the availability of suitable manually designed cost parameters \rahel{\( c^{M} \in \mathbb{R}^{b} \)};  
ii) \rahel{$c^{M}$} is extended and repeated across the prediction horizon (lines 1 and 2), resulting in \rahel{$C^{M}_{N_L}$ and $C^{M}_{N_S}$}, for the considered long and short-horizon MPC, respectively; iii) \rahel{$C^{M}_{N_S}$ is} used as a warm start for our learning algorithm. Therefore, our NN outputs \rahel{the correction values $\Delta C_{N_S}^{\theta}$}, which we add to the \rahel{warm start $C^{M}_{N_S}$} to infer the learned cost parameters \rahel{$C_{N_S}^{\theta}$ (lines 5 and 6)}; iv) Finally, we adopt the \rahel{mean squared error (MSE)} as an imitation loss between predicted and \rahel{long-horizon} states and control-input trajectories. \rahel{In practice, we consider only the first $N_D \leq N_S$ steps in the imitation loss, i.e., $MSE_{N_D}$ (see Appendix \ref{apx:learningframeworksetup} for more details.)}, and use Adam as the gradient-based optimizer to update the NN weights \rahel{(lines 7 to 10)}; v) Effectively, algorithm \ref{alg:ml_mpc} is executed in mini-batches, i.e., a batch of initial states is sampled, and the loss is computed using empirical risk minimization over the batch.

\section{Experiments}

Autonomous racing \rahel{poses} a safety-critical application in robotics, requiring fast computation times of the applied control action. Furthermore, \rahel{designing an MPC cost function that makes a car complete a lap in minimal time proves challenging for short-horizons, while convincing results have been shown with longer horizons (Appendix \ref{apx:laptime}).} 
This makes it a compelling application to demonstrate the usefulness of our proposed framework. Therefore, we specify the used system dynamics $f$ with two different models. The simpler \emph{kinematic bicycle}, and the more complex \emph{Pacejka} model \cite{rajamani2011vehicle}. Furthermore, representing the car state and race track in Frenet frame \cite{milliken1995race}, we design the controller as a \rahel{model predictive contouring controller (MPCC) \cite{liniger2015optimization} with a quadratic stage cost function $\ell$ and corresponding cost parameter vector $c_i = [q_i,p_i]$, where $q_i,p_i \in \mathbb{R}^{n+m}$. Accordingly, for later reference, we define $Q_{N} := \{q_i\}_{0}^{N}$ and $P_{N} := \{p_i\}_{0}^{N}$, as well as the corresponding parametric mapping functions $h_{Q}^{\theta}$ and $h_{p}^{\theta}$. Furthermore, we} chose the context variable to be \rahel{a long-horizon sequence of} track curvatures. Details on the car models and control approach are provided in  \mbox{Appendix \ref{apx:autonomousracing}}.

\paragraph{Experiment Design and Baselines.}
We evaluate our approach conducting experiments in simulation and real-world.
In Sec.~\ref{sec:main_results}, we assess the performance of ZipMPC by comparing it with BO and eMPC, \rahel{inspired by the approaches in \cite{Fröhlich2022} and \cite{quan2019approximate} and adapted to our framework}. We also consider two baselines that are given by a long-horizon MPC ($\mathit{MPC}_{N_L}$) with horizon length $N_L$, serving as a high-performance but time-consuming \rahel{baseline}, and a short-horizon MPC ($\mathit{MPC}_{N_S}$) with horizon length $N_S$, both employing \rahel{intuitively tuned} cost parameters \( q^{M} \) and \( p^{M} \) across the horizon. \rahel{Details on the choice of \( q^{M} \) and \( p^{M} \)  are provided in Appendix \ref{apx:modelparameters}}. To determine suitable values for the horizon lengths $N_S$ and $N_L$, we evaluate lap times with horizon lengths (see Tables \ref{tab:laptime} and \ref{tab:laptime_paj} in the Appendix) given the manually designed cost.
While ZipMPC follows Algorithm \ref{alg:ml_mpc} to learn a $\Delta$-cost that is context and state dependent, the BO baseline obtains a constant $\Delta$-cost using BO to minimize the imitation loss between $\mathit{MPC}_{N_S}$ and $\mathit{MPC}_{N_L}$. The eMPC baseline employs a fully learning-based control strategy, where a NN is employed to clone the behaviour of $\mathit{MPC}_{N_L}$. 
In Sec.~\ref{sec:main_results}, overall performances are measured based on trajectory imitation error and lap completion time, and we also demonstrate that our method generalizes well to unseen tracks. In Sec.~\ref{sec:ablation}, we analyze the role of the context in our approach and investigate the NN outputs. If not further specified, train and test track are chosen as illustrated in \mbox{Figure \ref{fig:trajs}}. Finally, in Sec.~\ref{sec:hardware} we test our proposed method on hardware. 

\subsection{Simulation Results}
\label{sec:main_results}

\paragraph{Imitation Error.} 
We trained an MPC cost function using the \emph{kinematic model} with various scenarios by varying the horizon lengths $N_S$ and $N_L$. For each scenario, the learned cost parameters ($Q_{N_S}^{\theta}$ and $P_{N_S}^{\theta}$), conditioned on the \rahel{sequence of} track curvatures and useful variables of the initial state \rahel{(for further details see Appendix \ref{apx:learningframeworksetup})}, were used to execute ZipMPC over a fixed set of 1000 validation initial states to obtain 1000 trajectories. Each trajectory was compared with the trajectory \rahel{resulting} from the reference method $\mathit{MPC}_{N_L}$ (root mean squared error (RMSE) between the states plus control variables). The obtained results are presented in Table \ref{tbl:main_im_loss}, showing an increased imitation capability of ZipMPC compared to the chosen baselines.

\begin{table*}[ht]
\centering
\small
\caption{Imitation loss (RMSE) between methods and $\mathit{MPC}_{N_L}$ of the first 5 steps using the \emph{kinematic model}. Mean and standard deviation are reported over a validation set of 1000 initial states.}
\begin{tabular}{cc|cccc}
\toprule
$N_S$ & $N_L$ & $\mathit{MPC}_{N_S}$ & BO & eMPC & ZipMPC  \\
\midrule
5 & 18 & $0.194 \pm 0.004$ & $0.089 \pm 0.009$ & $0.081 \pm 0.007$ & $\mathbf{0.065 \pm 0.006}$ \\
5 & 25 & $0.203 \pm 0.004$ & $0.150 \pm 0.006$ & $0.086 \pm 0.009$ & $\mathbf{0.068 \pm 0.009}$ \\
10 & 18 & $0.061 \pm 0.005$ & $0.061 \pm 0.005$ & $0.101 \pm 0.007$ & $\mathbf{0.043 \pm 0.016}$ \\
10 & 25 & $0.088 \pm 0.008$ & $0.086 \pm 0.008$ & $0.108 \pm 0.007$ & $\mathbf{0.043 \pm 0.012}$ \\
\bottomrule
\label{tbl:main_im_loss}
\end{tabular}
\vspace{-2.2em}
\end{table*}

\begin{figure*}[h!]
    \vspace{-0.5em}
    \centering
    \includegraphics[width=0.8\linewidth]{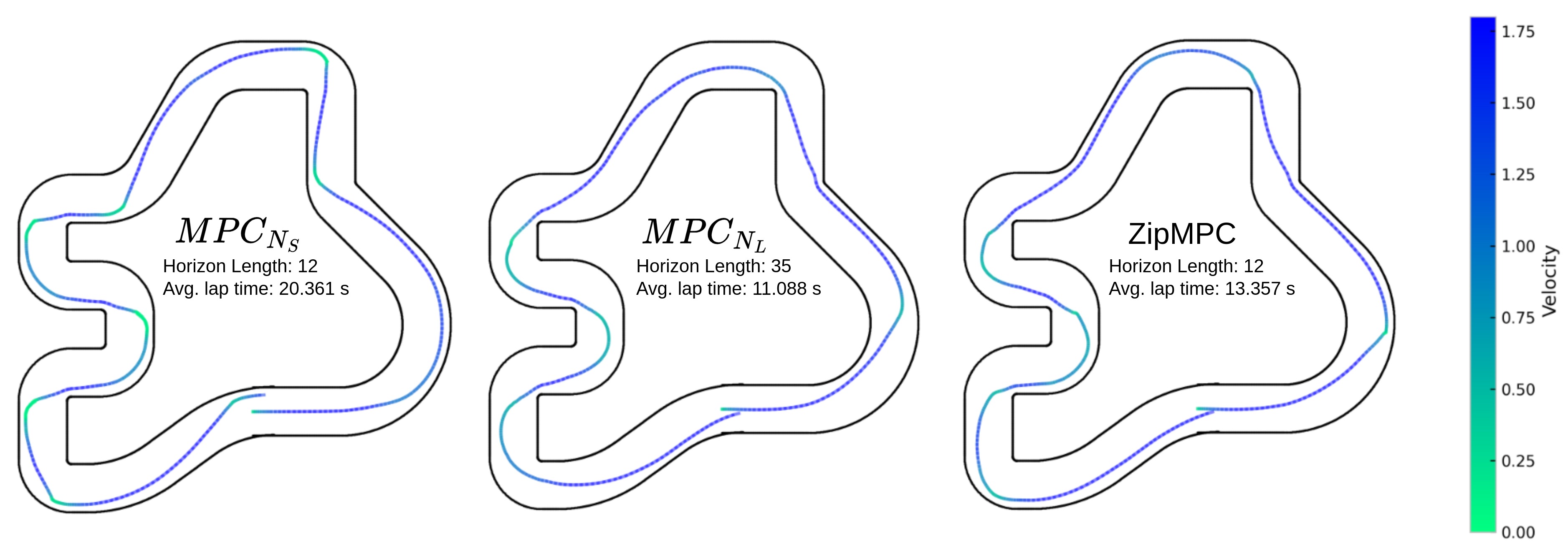}
    \vspace{-0.5em}
    \caption{Comparison of three closed-loop trajectories ($\mathit{MPC}_{N_S}$, $\mathit{MPC}_{N_L}$, and ZipMPC) with velocity information on the train track using the \emph{Pacejka model}. $\mathit{MPC}_{N_S}$ and $\mathit{MPC}_{N_L}$ correspond to the optimal trajectory given a manual cost set to the MPC with a horizon of $12$ and $35$ steps, respectively. Trajectories in this and all other figures are counterclockwise.}
    \label{fig:trajs}
    \vspace{-1.5em}
\end{figure*}

\paragraph{Lap Time.}
In Tables \ref{tbl:main_kin} and \ref{tbl:main_pac}, we demonstrate that ZipMPC considerably improves the lap time of $\mathit{MPC}_{N_S}$ in both the \emph{kinematics} and \emph{Pacejka model}, for all tested combinations of $N_S$ and $N_L$.
With the \emph{kinematics model} (Table \ref{tbl:main_kin}), our method achieves a lap time close to that of the reference method, $\mathit{MPC}_{N_L}$, and also outperforms the selected baseline models. Additionally, we provide an illustration of all three trajectories (from $\mathit{MPC}_{N_S}$, $\mathit{MPC}_{N_L}$, and ZipMPC) using the \emph{Pacejka} model in Figure \ref{fig:trajs}. The figure highlights the substantial improvement from $\mathit{MPC}_{N_S}$ to ZipMPC even though they have the same horizon length in the MPC optimization, as ZipMPC learns to smooth harsh curves present in $\mathit{MPC}_{N_S}$, similar to $\mathit{MPC}_{N_L}$, consequently reducing the average lap time. In Table \ref{tbl:main_pac}, we additionally present the optimization time reduction during the execution of ZipMPC compared to $\mathit{MPC}_{N_L}$. 

\begin{table*}[ht]
\centering
\small
\caption{Lap time (sec.) comparison using the \emph{kinematic model}. Mean and standard deviation are reported over ten runs with noise on the initial state. \mbox{Results marked with $-$ did not complete the lap.}}
\begin{tabular}{cc|c|cccc}
\toprule
$N_S$ & $N_L$ & $\mathit{MPC}_{N_L}$ (reference) & $\mathit{MPC}_{N_S}$ & BO & eMPC & ZipMPC \\
\midrule
5 & 18 & $8.358 \pm \scriptsize{0.015}$ & $9.103 \pm \scriptsize{0.043}$ & $8.847 \pm \scriptsize{0.016}$ & $-$ & $\mathbf{8.436 \pm \scriptsize{0.012}}$ \\
5 & 25 & $8.286 \pm \scriptsize{0.012}$ & $9.103 \pm \scriptsize{0.043}$ & $8.955 \pm \scriptsize{0.028}$ & $-$ & $\mathbf{8.394 \pm \scriptsize{0.012}}$ \\
10 & 18 & $8.358 \pm \scriptsize{0.015}$ & $8.556 \pm \scriptsize{0.018}$ & $8.580 \pm \scriptsize{0.013}$ & $-$ & $\mathbf{8.370 \pm \scriptsize{0.000}}$ \\
10 & 25 & $8.286 \pm \scriptsize{0.012}$ & $8.556 \pm \scriptsize{0.018}$ & $8.610 \pm \scriptsize{0.013}$ & $-$ & $\mathbf{8.325 \pm \scriptsize{0.015}}$ \\
\bottomrule
\label{tbl:main_kin}
\end{tabular}
\vspace{-2.2em}
\end{table*}

\begin{table}[ht]
\vspace{-1.0em}
\centering
\small
\caption{Lap time in seconds (left) and execution time reduction for each MPC step (right) using the \emph{Pacejka model}. Mean and standard deviation are reported over ten lap runs with slight noise on the initial state. Results marked with $-$ never completed a lap (infeasible). Results marked with $^*$ completed the lap at least 80\% of the attempts.}
\begin{tabular}{cc|c|cc||c}
\toprule
 & & \multicolumn{3}{c||}{Lap time (seconds)} & \multicolumn{1}{c}{Execution time reduction (per step)} \\
\midrule
$N_S$ & $N_L$ & $\mathit{MPC}_{N_L}$ (ref.) & $\mathit{MPC}_{N_S}$ & ZipMPC & $(1 - \frac{t(\text{ZipMPC})}{t(\mathit{MPC_{N_L}})})*100\%$ \\
\midrule
6 & 35 & $11.09 \pm \scriptsize{0.07}$ & $-$ & $\mathbf{13.83 \pm \scriptsize{0.43} ^*}$ & $82.5 \pm \scriptsize{1.1} \%$\\
6 & 45 & $9.89 \pm \scriptsize{0.10}$ & $-$ & $\mathbf{12.82 \pm \scriptsize{0.33}^*}$ & $87.9 \pm \scriptsize{0.4} \%$\\
12 & 35 & $11.09 \pm \scriptsize{0.07}$ & $20.36 \pm \scriptsize{0.20}$ & $\mathbf{13.36 \pm \scriptsize{0.47}}$ & $68.5 \pm \scriptsize{1.6}\%$\\
12 & 45 & $9.89 \pm \scriptsize{0.10}$ & $20.36 \pm \scriptsize{0.20}$ & $\mathbf{12.82 \pm \scriptsize{0.28}}$ & $76.6 \pm \scriptsize{0.5}\%$\\
\bottomrule
\label{tbl:main_pac}
\end{tabular}
\vspace{-2.2em}
\end{table}

\begin{figure*}[h]
    \centering
    \vspace{-0.5em}    \includegraphics[width=0.8\linewidth]{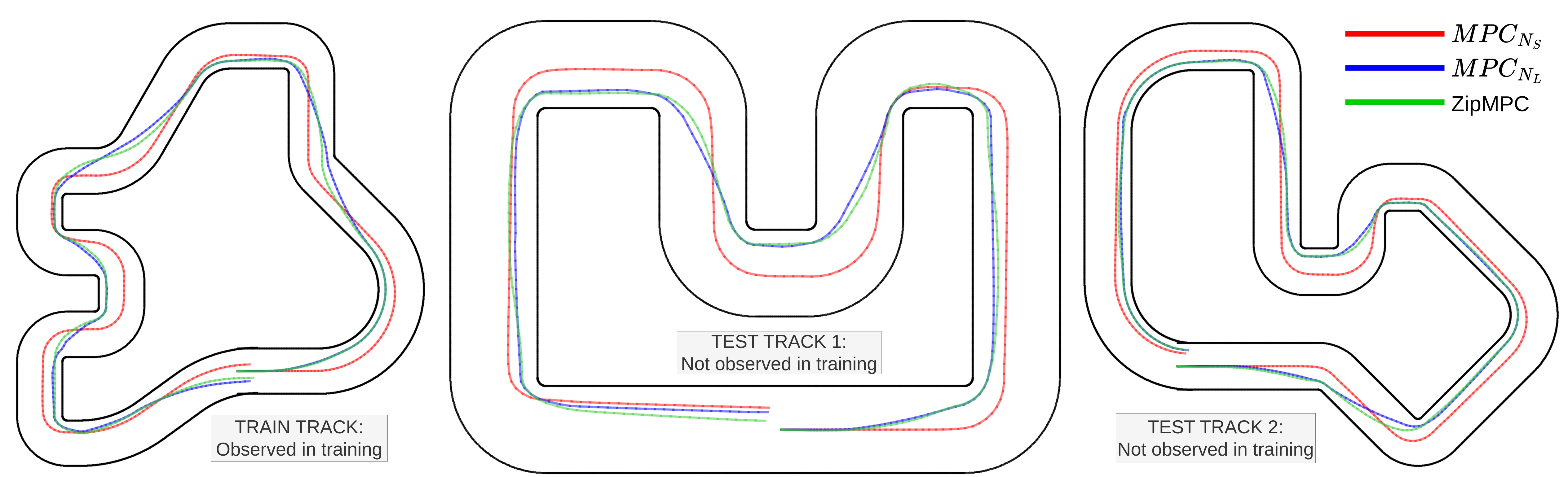}
    \caption{Trajectories from $\mathit{MPC}_{N_S}$, $\mathit{MPC}_{N_L}$, and ZipMPC on three tracks using the \emph{kinematics} model with $N_S=5$ and $N_L=25$. Only the left-most track was observed in the training process.}
    \label{fig:trajs_gen}
    \vspace{-1.0em}
\end{figure*}

\paragraph{Track Generalization.} To demonstrate the capability of our framework to generalize inferred trajectories across different tracks, we trained the model on a track that is complex enough to include a variety of curve sequences. We then tested the lap performance not only on the same track but also on two different tracks: one longer and one shorter, \rahel{containing curvatures of different steepness and directions, without extrapolating the maximum steepness}. The results, obtained using the \emph{kinematic} model, are presented in Table \ref{tbl:gen_kin_pacejka} and in Figure \ref{fig:trajs_gen}. In all scenarios, even in unseen tracks during training, ZipMPC was able to decrease the lap time compared to $\mathit{MPC}_{N_S}$, specifically using the \emph{kinematic} model, where the lap times of ZipMPC were very close to the target $\mathit{MPC}_{N_L}$.

\begin{table}[ht]
\centering
\small
\vspace{-1.1em}
\caption{Generalization \rahel{to} different tracks. Lap times (seconds) comparison are reported using both the \emph{kinematic} and the \emph{Pacejka} model executed on the trained and two new tracks. Mean and standard deviation are reported over ten runs with noise on the initial state.}
\label{tbl:gen_kin_pacejka}
\begin{tabular}{c|cc|c|c|cc}
\toprule
Dynamics model & $N_S$ & $N_L$ & Track & $\mathit{MPC}_{N_L}$ (reference) & $\mathit{MPC}_{N_S}$ & ZipMPC \\
\midrule
\multirow{3}{*}{Kinematics} & \multirow{3}{*}{5} & \multirow{3}{*}{25}
& Train   & $8.286 \pm \scriptsize{0.012}$ & $9.090 \pm \scriptsize{0.048}$ & $\mathbf{8.394 \pm \scriptsize{0.012}}$ \\
& & & Test 1  & $6.273 \pm \scriptsize{0.009}$ & $6.891 \pm \scriptsize{0.040}$ & $\mathbf{6.312 \pm \scriptsize{0.015}}$ \\
& & & Test 2  & $8.790 \pm \scriptsize{0.000}$ & $9.537 \pm \scriptsize{0.031}$ & $\mathbf{8.823 \pm \scriptsize{0.009}}$ \\
\midrule
\multirow{3}{*}{Pacejka} & \multirow{3}{*}{12} & \multirow{3}{*}{45}
& Train   & $9.894 \pm \scriptsize{0.100}$ & $20.361 \pm \scriptsize{0.201}$ & $\mathbf{12.822 \pm \scriptsize{0.283}}$ \\
& & & Test 1  & $8.718 \pm \scriptsize{0.034}$ & $15.123 \pm \scriptsize{0.076}$ & $\mathbf{12.000 \pm \scriptsize{0.075}}$ \\
& & & Test 2  & $10.245 \pm \scriptsize{0.040}$ & $17.736 \pm \scriptsize{0.439}$ & $\mathbf{15.123 \pm \scriptsize{0.078}}$ \\
\bottomrule
\end{tabular}
\vspace{-1.0em}
\end{table}

\subsection{Ablation Study in Simulation}
\label{sec:ablation}

\begin{figure}[h]
    \vspace{-1.5em}
    \centering
    \begin{subfigure}[h]{0.38\textwidth}
        \centering
        \includegraphics[height=1.3in]{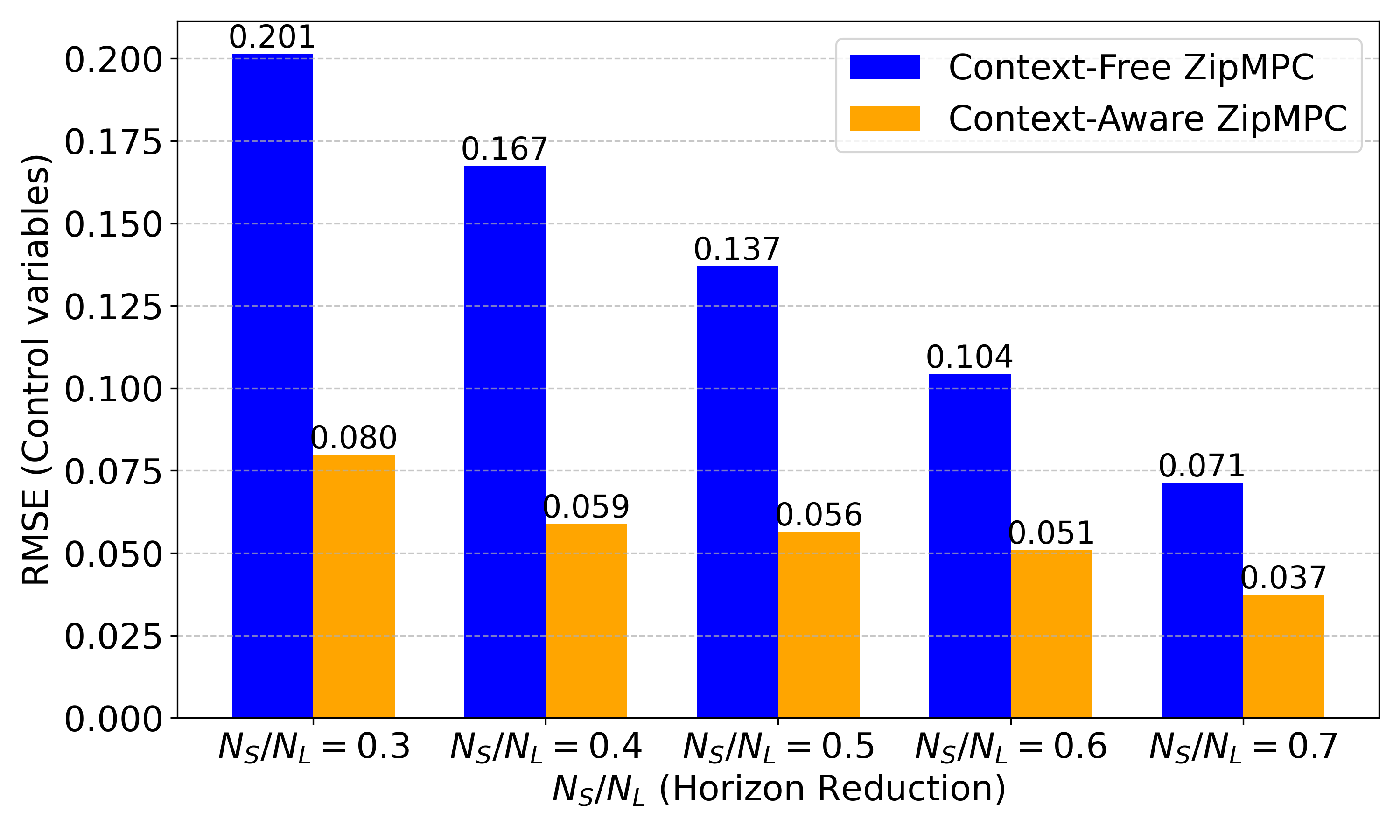}
        \caption{RMSE comparison of context-free and aware ZipMPC across different ratios of short (learned) and long-horizons ($N_S/N_L$). $N_L=20$ was set in this experiment for the \emph{kinematic} model.}
        \label{fig:context_on_rmse}
    \end{subfigure}
    ~    
    \begin{subfigure}[h]{0.58\textwidth}
        \centering
        \includegraphics[height=1.3in]{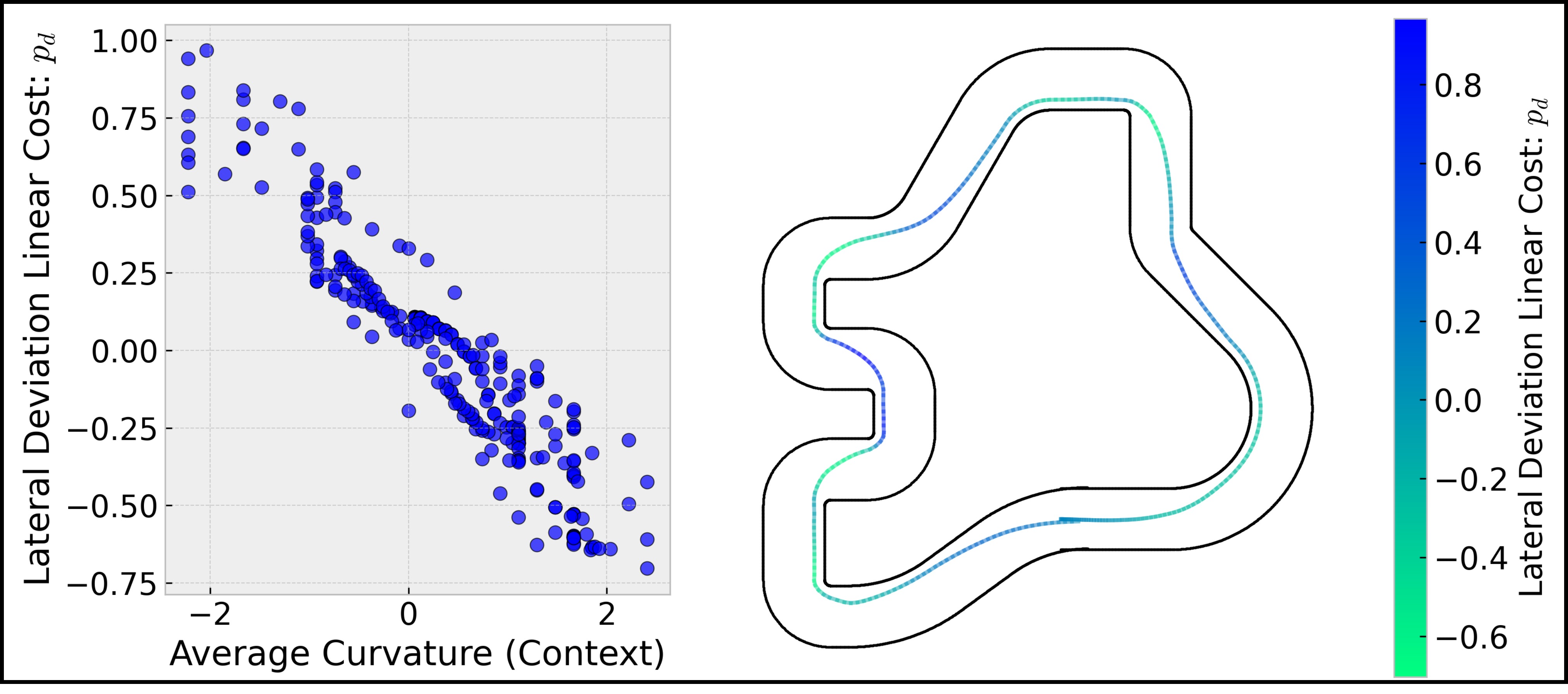}
        \caption{The left graph illustrates the correlation between \( p_d \) and the average curvature observed over the prediction horizon. The right graph shows the value of the learned parameter \( p_d \) along the trajectory. The experiment was using $N_S=5$ and $N_L=18$.}
        \label{fig:context_on_p}
    \end{subfigure}%
    \vspace{-0.3em}
    \caption{Ablation experiments for the \emph{kinematics bicycle model}}
    \label{fig:ablation}
    \vspace{-1.5em}
\end{figure}

\paragraph{Context relevance.}
We investigate the importance of incorporating context as input to the NN in our framework. To this end, we compare our proposed method, where the NN optimizes $h_{Q}^{\theta}(x(k), Z_{k,N_L})$ and $h_{P}^{\theta}(x(k), Z_{k,N_L})$, against a baseline variant of the framework where the context is excluded, and the NN infers $h_{Q}^{\theta}(x(k))$ and $h_{P}^{\theta}(x(k))$. The results, provided in Figure \ref{fig:ablation}-a, indicate that for shorter prediction horizons (\(N_S\)), incorporating context from the long-horizon (\(N_L\)) provides a substantial performance improvement. This benefit diminishes as \(N_S\) approaches \(N_L\), where the context difference becomes less critical for the learning model. Furthermore, we observe a trend of improved performance with increasing \(N_S\), attributable to better initial estimates of the MPC cost and enhanced capacity of the MPC cost optimization.

\paragraph{Learned Parameters}  
To better understand the role of context in improving trajectory generation, we analyze how the learned cost component associated with lateral deviation (denoted here as \( p_d \), i.e., the linear cost corresponding to the lateral deviation in \( P_{N_S}^{\theta} \)) varies along the track. As shown in Figure \ref{fig:ablation}-b, the graph on the right reveals significant variability in the \( p_d \) values throughout the track. Specifically, right-hand curves result in learned positive \( p_d \) values, while left-hand curves lead to learned negative \( p_d \) values, aligning with intuitive expectations. The left graph displays the same \( p_d \) values in a scatter plot, plotted against the curvature. This plot highlights a strong correlation between the average observed upcoming curvature and the learned \( p_d \) values.

\subsection{\rahel{Real World} Results}
\label{sec:hardware}

\begin{wrapfigure}{r}{0.30\linewidth}
\vspace{-1.5em}
    \centering
    \includegraphics[width=\linewidth]{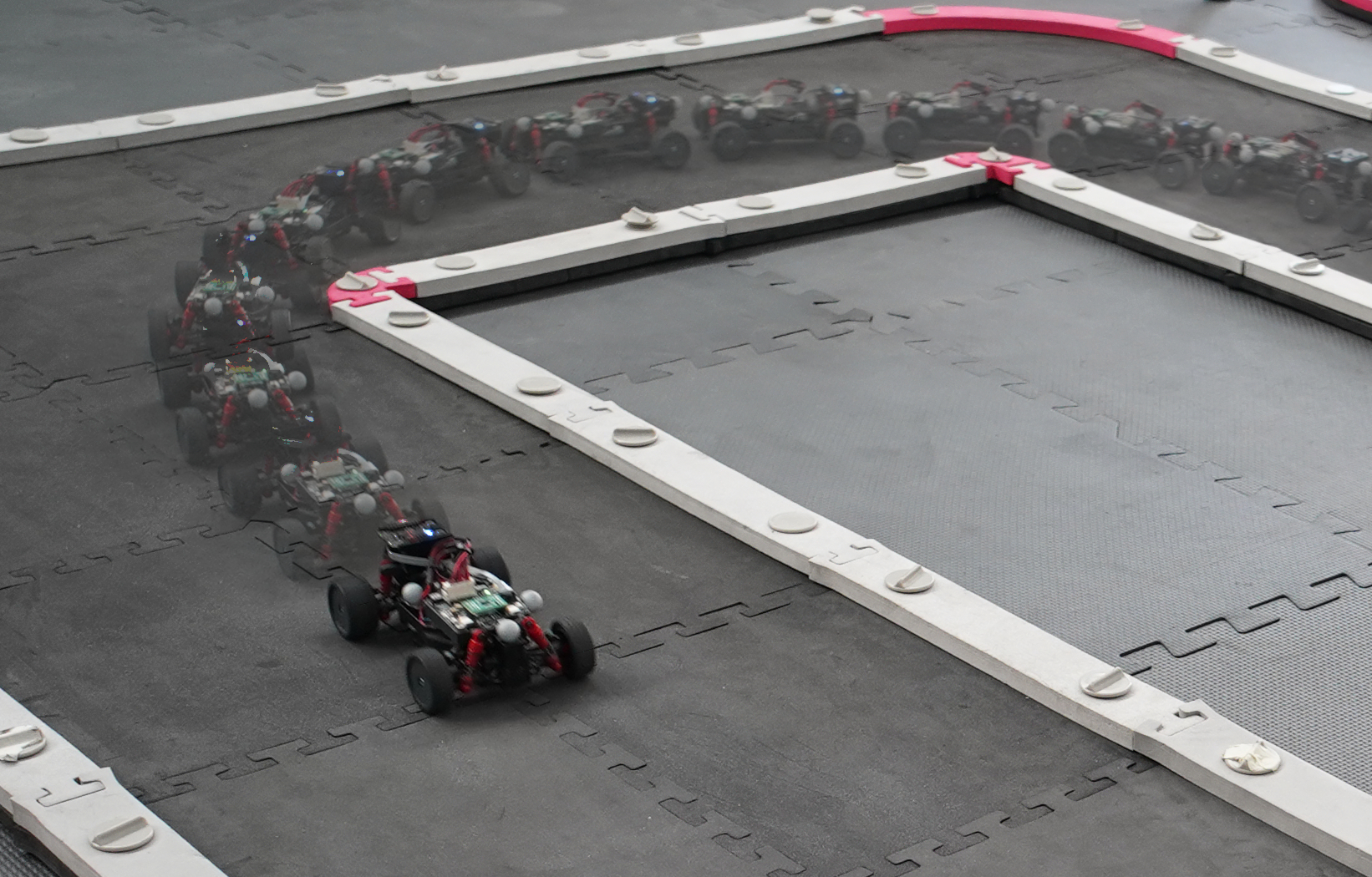}
    \caption{Illustration of the employed miniature racing platform.}
    \label{fig:pic_car}
\end{wrapfigure}
\begin{figure}[h]
    \vspace{-1.0em}
    \centering
    \includegraphics[width=\linewidth]{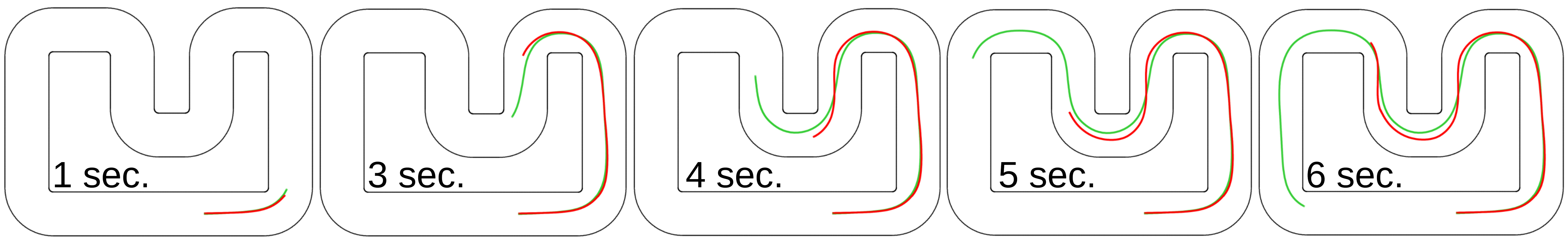}
    \caption{Comparison of \rahel{closed-loop trajectories of} ZipMPC (green) and $\mathit{MPC}_{N_S}$ (red) for different instances in time, executed on hardware. Thereby $N_S$ is chosen as 20 and $N_L$ is set equal to 40.}
    \label{fig:hardware_20_40}
    \vspace{-1em}
\end{figure}

Using the test bench presented in \cite{carron2023chronos} and illustrated in Figure \ref{fig:pic_car}, we verified our simulation results on hardware. Using the \textit{Pacejka} model, we trained the ZipMPC for various combinations of $N_S$ and $N_L$ in simulation and employed the obtained model on hardware without any retraining. Figure \ref{fig:hardware_20_40} illustrates a comparison of \rahel{the closed-loop trajectories of} our proposed ZipMPC and $\mathit{MPC}_{N_S}$ with $N_S=20$ and $N_L=40$ for different instances in time. It can be seen how the ZipMPC incorporates the additional contextual information to improve racing performance. Results for further horizon lengths are presented in Appendix \ref{apx:additionalhardwareresults}. \textbf{Accompanying videos are found in the supplementary material.}

\section{Conclusion}
Employing differentiable model predictive control (MPC) and neural networks, we developed a new algorithm to imitate long-horizon MPC solutions using an MPC with a significantly shorter prediction horizon. This is achieved by incorporating long-horizon contextual information into a \rahel{learned} cost vector for the short-horizon MPC. \rahel{The conducted experiments show} that the proposed method combines the strengths of explicit model predictive control with the tools for autonomous tuning of cost parameters. This includes a reduction of the computational time required to optimize controller inputs, the preservation of constraint feasibility with respect to system dynamics, and enhanced generalization capabilities to unseen environments \rahel{during training}. The conducted \rahel{real world} experiments on a miniature racing platform further \rahel{validate} the observations from the simulation.

\newpage
\section{Limitations}
In the following section, we discuss on current limitations of the proposed method, as well as potential strategies on how to address them.  
\paragraph{Dependency on differentiable MPC method.}
The performance of a resulting ZipMPC is highly dependent on the capabilities of the chosen method to propagate gradients through the MPC optimization process, and therefore directly \rahel{inherits} its limitations. In the presented case, this includes inaccuracies resulting from the convex approximations at each time step, as well as the usage of exact penalties to allow for the inclusion of inequality constraints on state and not only input variables. However, these limitations are expected to {be reduced} with future developments in the corresponding field of research. 
\paragraph{Sentivity to infeasible training samples.}
Furthermore, our method has shown sensitive behaviour to the sampling of infeasible initial states during training, i.e., states for which the solver is not capable of finding a solution that fulfills the existing constraints for all time steps of the chosen prediction horizon. While finding an approximation of the feasible region was not too difficult for the problem description at hand, this might be harder for more complex problem descriptions. Nevertheless, such an issue can be addressed by running a standard solver in parallel, throwing an error whenever an infeasible initial state is detected, and neglecting the corresponding sampling point for training. 
\paragraph{Nonconvexity of imitation loss.} 
The nonconvexity of the proposed imitation loss with respect to the parameters $\theta$ of our parametric mapping function might result in convergence to highly suboptimal local minima. While we did not experience any nonconvexity issues employing the \textit{kinematic} model, we addressed them for the \textit{Pacejka} model by deploying a selection mechanism to save the model with the best lap performance in a fixed number of iterations. 
\paragraph{Theoretical guarantees.}
Our proposed method is partly motivated by the results on long-horizon MPC stability and recursive feasibility without terminal ingredients. However, given the slightly imperfect approximation of $\mathit{MPC}_{N_L}$ with ZipMPC, we cannot assume to inherit the provided theoretical guarantees of the former but only show them empirically. Investigations of this matter, e.g., quantifying the approximation error, present \rahel{themselves} as an interesting topic for future work.

\clearpage
\acknowledgments{This work has been partially supported by the Industrial Graduate School Collaborative AI \& Robotics funded by the Swedish Knowledge Foundation Dnr:20190128, and the Knut and Alice Wallenberg Foundation through Wallenberg AI, Autonomous Systems and Software Program (WASP).}


\bibliography{main}  

\clearpage

\appendix

\section{Autonomous Racing as an Application Example}
\label{apx:autonomousracing}
In this section, we detail the preliminaries for modeling an autonomous racing application. It includes the Frenet coordinate frame in \ref{apx:frenetframe}, the corresponding car dynamics in \ref{apx:cardynamics}, as well as the chosen MPC design in \ref{apx:mpcforracing}. Furthermore, we comment on the employed model and control parameter values in \ref{apx:modelparameters}. 

\subsection{Frenet Frame}
\label{apx:frenetframe}
The Frenet coordinate frame \cite{milliken1995race} has emerged as one of the standard coordinate frames in racing applications. It describes a car's position with respect to a given trajectory using its progress along the trajectory, its lateral deviation from the trajectory, and its orientation with respect to the trajectory, indicated with $\sigma$, $d$, and $\phi$, respectively. We denote the corresponding state vector as $x = [\sigma,d,\phi]$ and illustrate it in Figure \ref{fig:frenetkin}.

\subsection{Car Dynamics}
\label{apx:cardynamics}
\begin{figure}[htb]
\vspace{-0.5em}
    \centering
    \includegraphics[width=0.3\textwidth]{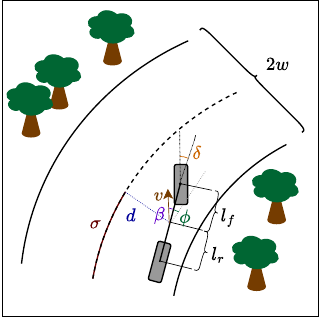}
    \caption{Illustration of the state vector defining a car's position in the Frenet frame, including its progress coordinate $\sigma$, its lateral deviation $d$, and its respective orientation $\phi$. Furthermore indicated are its velocity $v$ and the steering angle $\delta$ as well as the cars rear and front length, $l_r$ and $l_f$.}
    \label{fig:frenetkin}
    \vspace{-1.0em}
\end{figure}

To model the car's dynamics, we rely on the \textit{kinematic bicycle} and the \textit{Pacejka} model \cite{rajamani2011vehicle,brunke2020learning}, both of which are further detailed below. 
For this purpose, we denote the car's rear length with $l_r$ and the distance from the center of mass to its front wheel axis with $l_f$. Additionally, we introduce the progress-dependent curvature model $\kappa(\sigma)$, which later on defines our context variable $\zeta$. All considered tracks are modeled with constant track width $2\omega$. For an illustration of the introduced parameters, you can once again refer to Figure \ref{fig:frenetkin}.

\subsubsection{Pacejka Model used in Simulation}
The \textit{Pacejka} model is primarily suitable for modeling high-velocity applications but is prone to numerical difficulties when dealing with small velocities. It takes a torque $\tau$ and a steering angle of the front wheels $\delta$ as control inputs and models the states $\sigma$, $d$, $\phi$, $r$, $v_x$, and $v_y$, with the latter defining yaw-rate, longitudinal and lateral velocity, respectively. Then, the lateral slip angles are calculated as 
\begin{align*}
    & \alpha_{f}(k) = - \arctan2\left(\frac{-v_y(k)- l_f r(k)}{\vert v_x(k) \vert} \right)+\delta(k), \quad \alpha_{r}(k) = - \arctan2\left(\frac{-v_y(k) + l_f r(k)}{\vert v_x(k) \vert}\right),
\end{align*}
and the lateral tire forces $F_{f}$ and $F_{r}$, as well as the motor force $F_{m}$ follow as 
\begin{align*}
    & F_{f}(k) = -D_f \sin(C_f \tan^{-1}(B_f a_f(k))) \\
    & F_{r}(k) = -D_r \sin(C_r \tan^{-1}(B_r a_r(k))) \\
    & F_{m}(k) = (C_{m1} - C_{m2}   v_{x}(k))   \tau - C_d   v_x(k)   v_x(k) - C_{roll}
\end{align*}
with $D_f$, $C_f$, $B_f$, $D_r$, $C_r$, and $B_r$ describing the lateral force parameters of front and rear wheels, as well as $C_{m1}$, $C_{m2}$, $C_d$ and $C_{roll}$ defining the longitudinal force parameters. Additionally, denoting with $m$ the mass of the car, with $I_z$ its moment of inertia, and with $T$ the chosen discretization time, the discrete-time dynamics, discretized using the Euler method, follow as
\begin{align*}
    &\sigma(k+1) = \sigma(k) + T\left( \frac{v_{x}(k)\cos(\phi(k))-v_{y}(k) \sin(\phi(k))}{1-\kappa(\sigma(k))d(k)} \right) \\
    &d(k+1) = d(k) + T\left(v_{x}(k) \cos(\phi(k)) + v_{y}(k) \sin(\phi(k)) \right) \\
    &\phi(k+1) = \phi(k) + T\left( r(k)-\kappa(\sigma(k))\frac{v_{x}(k) \cos(\phi(k))-v_{y}(k) \sin(\phi(k))}{1-\kappa(\sigma(k))d(k)} \right)\\
    &r(k+1) = r(k) + T\left(\frac{1}{I_z}(F_f(k)  l_f  \cos(\delta(k)) - F_r(k)  l_r) \right) \\
    &v_{x}(k+1) = v_{x}(k) + T\left( \frac{1}{m}(F_m(k) - F_f(k) \sin(\delta(k)) + m  v_{y}(k)  r(k)) \right) \\
    &v_{y}(k+1) = v_{y}(k) + T\left( \frac{1}{m}(F_r(k) + F_f(k) \cos(\delta(k)) - m  v_{x}(k)  r(k)) \right).
\end{align*}

\subsubsection{Pacejka Model used on Hardware}
For our obtained real world results we employed a slightly adapted version of the \textit{Pacejka} model used in simulation to increase the model accuracy. 
Therefore, we additionally define the friction force $F_{fr}$, which, introducing the longitudinal parameters $C_{d0}$, $C_{d1}$ and $C_{d2}$, follow as
\begin{align*}
    & F_{fr}(k) = \mathrm{sign}(v_x(k)) (-C_{d0} - C_{d1} v_x(k) - C_{d2} v_x(k) v_x(k)).
\end{align*}
Furthermore, we alter the motor force $F_{m}$ and split $F_f$ and $F_r$ into $x$ and $y$ components, obtaining $F_{fx}$, $F_{fy}$ and $F_{rx}$, $F_{ry}$, respectively. With $\gamma$ denoting the ratio between front and rear wheel drive force, these modifications result in 
\begin{align*}
    & F_{m}(k) = (C_{m1} - C_{m2}   v_{x}(k))   \tau \\
    & F_{fx}(k) = F_{m}(k) (1 - \gamma) \\
    & F_{rx}(k) = F_{m}(k) \gamma \\ 
    & F_{fy}(k) = D_f \sin(C_f \tan^{-1} (B_f a_f(k))) \\
    & F_{ry}(k) = D_r \sin(C_r \tan^{-1} (B_r a_r(k))).
\end{align*}
Correspondingly, these changes cause a different one-step forward prediction of $r$, $v_x$, and $v_y$, detailed in the following. 
\begin{align*}
    & r(k+1) = r(k) + T\left(\frac{1}{I_z}(F_{fy}(k)  l_f  \cos(\delta(k)) + F_{fx}(k)  l_f  \sin(\delta(k)) - F_{ry}(k)  l_r) \right) \\
    & v_{x}(k+1) = v_{x}(k) + T\left( \frac{1}{m}(F_m(k) - F_{fy}(k) \sin(\delta(k)) + F_{fx}(k) \cos(\delta(k)) + m  v_{y}(k)  r(k)) + F_{fr} \right)  \\
    & v_{y}(k+1) = v_{y}(k) + T\left( \frac{1}{m}(F_{ry}(k) + F_{fy}(k) \cos(\delta(k)) + F_{fx}(k) \sin(\delta(k)) - m  v_{x}(k)  r(k)) \right).
\end{align*}

\subsubsection{Kinematic Bicycle Model}
Differently from the \textit{Pacejka} model, the \textit{kinematic bicycle} model does not model tire forces or lateral velocities, but has shown its usefulness primarily in small to medium-velocity applications. Furthermore, it is computationally cheaper in its application than the \textit{Pacejka} model.
The \textit{kinematic bicycle} model takes the acceleration $a$, as well as the front wheel steering angle $\delta$ as control inputs and defines the side slip angle 
\begin{align*}
    \beta(k) = \tan^{-1}(\frac{l_r}{l_f+l_r}\tan(\delta(k))).
\end{align*}

The discrete-time dynamics, discretized using the Euler method, accordingly follow as
\begin{align*}
    &\sigma(k+1) =  \sigma(k) + T(v(k)\frac{\cos{(\phi(k) + \beta(k))}}{1 - \kappa(\sigma(k))d(k)}) \\
    &d(k+1) = d(k) +  T(v(k)\sin{\phi(k) + \beta(k)}) \\
    &\phi(k+1) = \phi(k) +  T(\frac{v(k)}{l_f}\sin{(\beta(k))} - \kappa(\sigma(k))v(k) \frac{\cos{(\phi(k)+\beta(k))}}{1 - \kappa(\sigma(k))d(k)})\\
    & v(k+1) = v(k) + Ta(k).
\end{align*}

\subsection{MPC for Racing}
\label{apx:mpcforracing}
In the Frenet frame, the classical Model
Predictive Contouring Controller (MPCC), one of the most commonly used planning and control algorithms for autonomous racing in the Cartesian frame \cite{liniger2015optimization}, can be represented as a nonlinear MPC controller with quadratic cost and polytopic constraint. The latter holds since the track boundary constraint transforms into a two-sided half-space constraint for a constant track width $2\omega$. Additional limitations on state and control-input variables can be implemented via the polytopic inequality constraint $G[x_i,u_i]^{\top} \leq g$, where $G \in \mathbb{R}^{c \times (n+m)}$ and $g \in \mathbb{R}^{c}$. Accordingly, the optimization problem, solved in receding horizon fashion, follows as 
\begin{align*}
    [X^*_{N+1}, U^*_{N}] := \argmin_{\{x_i\}_{0}^{N+1}, \{u_i\}_{0}^{N}} & \sum_{i=0}^{N} \Vert [x_i, u_i]\Vert_{q_i \mathbb{I}}^2 + \langle p_i, [x_i, u_i] \rangle\\
    \mathrm{s.t.} \ \ & x_{i+1} = f(x_i,u_i, \kappa(\sigma_i)), \\
    &  -\omega \leq d_i \leq \omega, \\
    & G[x_i,u_i]^{\top} \leq g, \\
    & x_0 = x(k), \\ 
    & i=0,\hdots,N.
\end{align*}
Note that depending on the choice of the car dynamics model, $f(x,u)$ is either replaced with the \textit{Pacejka} or the \textit{kinematic} model. Furthermore, to avoid cost scaling with an increasing progress value $\sigma_i$, we introduce an additional state $\sigma_{\Delta,i} = \sigma_i - \sigma_0$, which is independent of the current position on the race track, but only reflects the progress within the current prediction horizon. Finally, to increase clarity, we already indicated the context dependence, i.e., the dependence on track curvature $\kappa(\sigma)$, in the dynamics.   

\subsection{Model Parameters}
\label{apx:modelparameters}
In this subsection, we detail the deployed model, constraint, and objective parameter values for the presented experiments in simulation and hardware, respectively. For the experiments conducted in simulation with the \emph{kinematic} model, the manual cost vectors $q^{M}$ and $p^{M}$ are chosen identical for the long and short-horizon and are presented in Table \ref{tab:kin_costs}. For the experiments conducted in simulation with the $\emph{Pacejka}$ model, the manual cost vectors $q^{M}$ and $p^{M}$ are chosen as presented in Table \ref{tab:sim_costs}. Finally, for the experiments conducted on hardware, the manual cost vectors $q^{M}$ and $p^{M}$ are chosen dependent on the deployed horizon length and follow as presented in Table \ref{tab:hw_costs}.

\begin{table}[h]
\centering
\vspace{-1.0em}
\caption{Manual cost vectors used in simulation with the \emph{kinematic} model.}
\label{tab:kin_costs}
\begin{tabular}{lcccccccc}
\toprule
Parameter & $\sigma$ & $d$ & $\phi$ & $v$ & $\sigma_0$ & $\sigma_{\Delta}$ & $a$ & $\delta$ \\
\midrule
$q^{M}=$ & 0 & 3 & 1 & 0.01 & 0.01 & 0.01 & 0.01 & 1 \\
$p^{M}=$ & 0 & 0 & 0 & 0    & 0    & -8   & 0    & 0 \\
\bottomrule
\end{tabular}
\end{table}

\begin{table}[h]
\centering
\vspace{-1.0em}
\caption{Manual cost vectors used in simulation with the \emph{Pacejka} model}
\label{tab:sim_costs}
\begin{tabular}{lcccccccccc}
\toprule
Parameter & $\sigma$ & $d$ & $\phi$ & $r$ & $v_x$ & $v_y$ & $\sigma_0$ & $\sigma_{\Delta}$ & $\tau$ & $\delta$ \\
\midrule
$q^{M}=$ & 0 & 50 & 0.1 & 0.1 & 0.1 & 0.1 & 0.1 & 0.1 & 0.1 & 0.1 \\
$p^{M}=$ & 0 & 0  & 0   & 0   & 0   & 0   & 0   & -8  & 0   & 0   \\
\bottomrule
\end{tabular}
\end{table}

\begin{table}[h]
\centering
\vspace{-1.0em}
\caption{Manual cost vectors used on the \emph{Pacejka} model for hardware (scaled by $N$)}
\label{tab:hw_costs}
\begin{tabular}{lcccccccccc}
\toprule
Parameter & $\sigma$ & $d$ & $\phi$ & $r$ & $v_x$ & $v_y$ & $\sigma_0$ & $\sigma_{\Delta}$ & $\tau$ & $\delta$ \\
\midrule
$N*q^{M} =$ & 0 & 500 & 5 & 1 & 1 & 1 & 1 & 1 & 1 & 100 \\
$N*p^{M} =$ & 0 & 0 & 0 & 0 & 0 & 0 & 0 & -40 & 0 & 0 \\
\bottomrule
\end{tabular}
\end{table}

The model and constraint parameter values employed for the considered models in simulation and on hardware are presented in Table \ref{tbl:model_params}. 

\begin{table*}[h!]
\centering
\vspace{-0.5em}
\small
\caption{Model and constraint parameter values employed for the \textit{kinematic} model in simulation, the \textit{Pacejka} model in simulation, as well as the \textit{Pacejka} model on hardware.}
\begin{tabular}{c|ccc}
\toprule
Parameter & Kinematic (sim) & Pacejka (sim) & Pacejka (hard) \\
\midrule
$l_r$ & $0.05$ & $0.05$ & $0.038$ \\
$l_f$ & $0.05$ & $0.05$ & $0.052$\\
$m$ & $-$ & $0.200$ & $0.181$\\
$T$ & $0.03$ & $0.03$ & $0.026$\\
$\omega$ & $0.2$ & $0.2$ & $0.2$\\
$D_f$ & $-$ & $0.43$ & $0.65$\\
$C_f$ & $-$ & $1.4$ & $1.5$\\
$B_f$ & $-$ & $0.5$ & $5.2$\\
$D_r$ & $-$ & $0.6$ & $1.0$\\
$C_r$ & $-$ & $1.7$ & $1.45$\\
$B_r$ & $-$ & $0.5$ & $8.5$\\
$C_{m1}$ & $-$ & $0.9803$ & $0.9803$\\
$C_{m2}$ & $-$ & $0.0181$ & $0.0181$\\
$C_d0$ & $-$ & $0.0275$ & $0.085$\\
$C_d1$ & $-$ & $-$ & $0.01$\\
$C_d2$ & $-$ & $-$ & $0.0275$\\
$C_{roll}$ & $-$ & $0.085$ & $-$\\
$a_{\max}$ & $1.0$ & $1.0$ & $1.0$\\
$\delta_{\max}$ & $0.4$ & $0.5$ & $0.4$\\
$v_{\max}$ & $1.8$ & $1.8$ & $2.0$\\
\bottomrule
\end{tabular}
\label{tbl:model_params}
\end{table*}

\section{Experiment Setup}
In the following, we detail the learning framework in \ref{apx:learningframeworksetup} and provide some further information on the chosen hardware setup in \ref{apx:hardwaresetup}. In \ref{apx:setuplimitations} we discuss some limitations of the selected setup. 

\subsection{Learning Framework Setup and Implementation Details}
\label{apx:learningframeworksetup}

\paragraph{Input to the Neural Network.}
As we described in the main paper, the input of the NN is composed by $Z_{k, N_L}$ and some state variables from $x(k)$. $Z_{k, N_L}$ is composed by a timeseries of the track curvature from the current state $x(k)$. Specifically, we pick as many curvature values as needed to reach the maximum possible \rahel{$\sigma_{\Delta}$} in the long-horizon MPC optimization. \rahel{For example, with a discretization time $T=0.03$, a horizon length $N_L = 40$, and a maximum velocity $v_{max} = 2m/s$, we calculate $\sigma_{\Delta} = 0.03*40*2 = 2.4$, and pick all curvature points available between $\kappa(\sigma(k))$ and $\kappa(\sigma(k) + 2.4)$ to fill the $Z_{k, N_L}$ vector.}
Regarding the state variables from $x(k)$, we only choose the variables $v$, $d$, and $\phi$ because these state variables are general enough and useful to be applied to other tracks. This is not true for $\sigma$, for example, which would lead to overfitting to the specific training track.

\paragraph{Neural Network and Training.}
The neural network architecture used to predict the parameters of the MPC cost is composed of a convolutional feature extractor followed by fully connected layers. The model takes as input a concatenation of three global context variables represented by $v$, $d$, and $\phi$ coming from the initial state, and a time series $Z_{k, N_L}$ with a horizon of \( N_L \) representing the next curvature steps. We process the data through convolutional layers, batch normalization and dropout layers. We pass this shared representation through four fully connected layers, each with 512 hidden units and LeakyReLU activation. The final shared representation is used to compute two outputs: (1) a global representation layer producing a fixed component of the MPC cost, and (2) a modulation layer generating time-varying components. The modulation layer produces outputs with dimensions \( 2 \times N_S \times (n + m) \), i.e., we learn both $P^{\theta}_{N_S}$ and $Q^{\theta}_{N_S}$. The final output of the model is scaled to ensure the predicted parameters are between a certain range. Further details can be found in the code provided in the supplementary material.

\paragraph{Imitation Loss and Validation.}
As detailed in the main text, the imitation loss we implement is the sum of the MSE for the state variables and the MSE for the control variables. Each of these components are computed between the inferred variables from the ZipMPC and $\mathit{MPC}_{N_L}$. However, the range of different variables varies. Therefore, we manually weigh the different variables to avoid having one or more variables that are much more important to the loss function than others. The weights can be found in the code provided in the supplementary material. Also, we consider only the first $N_D$ steps of the trajectories from ZipMPC and $\mathit{MPC}_{N_L}$ in the imitation loss for two main reasons. The first is to make the loss function less complicated, such that it reaches a better local minimum. The second is that we have observed that the first trajectory steps are the most important to be aligned, which is also in line with the receding horizon implementation of MPC, where only the first optimal control input (the most imminent one) is applied to the system. For validation purposes, we compute lap performance (based on a fixed initial state) using the training track after a fixed number of training iterations. After a large number of iterations, we select the model with the best lap performance on that fixed initial state.

\subsection{Hardware Setup}
\label{apx:hardwaresetup}
\begin{figure}[h]
\vspace{-0.5em}
    \centering
    \begin{subfigure}[h]{0.48\textwidth}
        \centering
        \includegraphics[height=1.3in]{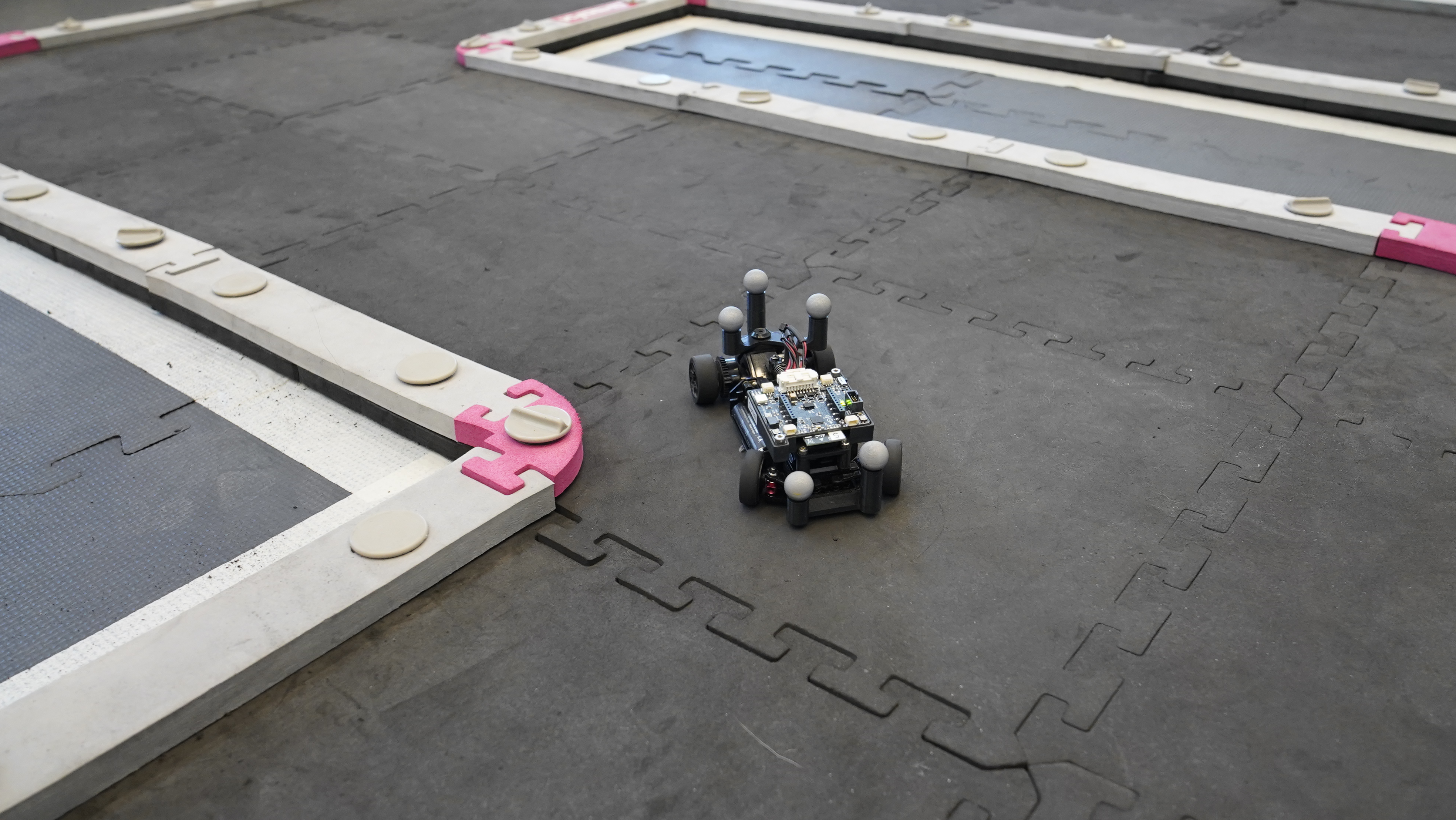}
        \caption{Picture of the employed miniature race car on race track.}
        \label{fig:car}
    \end{subfigure}
    ~    
    \begin{subfigure}[h]{0.48\textwidth}
        \centering
        \includegraphics[height=1.3in]{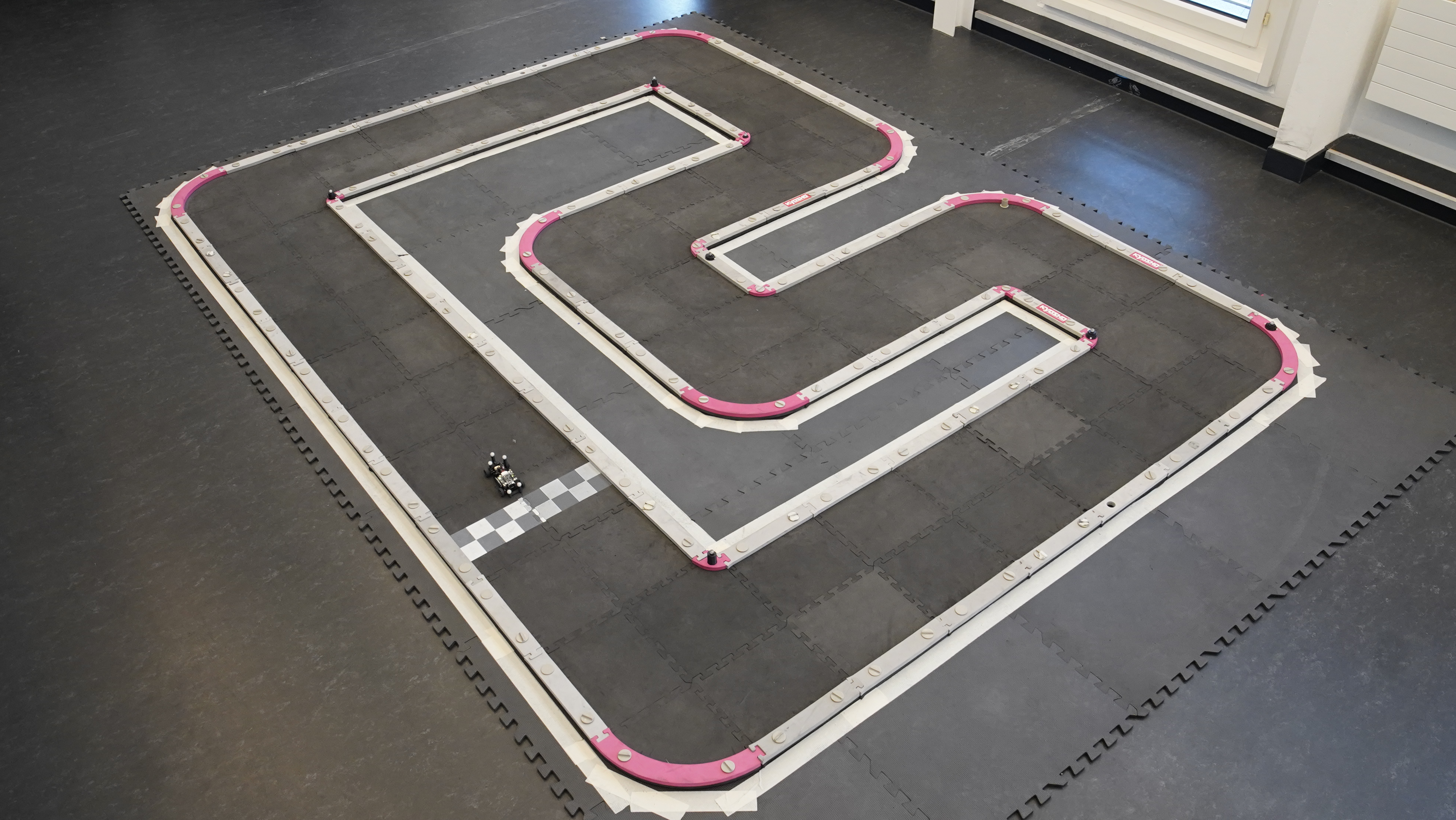}
        \caption{Picture of the race track on which the real world results are obtained. }
        \label{fig:fulltrack}
    \end{subfigure}%
    \caption{Pictures of used hardware setup.}
    \label{fig:figcrs}
    \vspace{-1.5em}
\end{figure}

For the obtained hardware results we employed the setup presented in \cite{carron2023chronos}. There, the position and orientation of a 1/28-scaled miniature car (Figure \ref{fig:car}) is measured using a motion capture system, passing the information to a server that runs the control algorithm. The car is connected to the aforementioned server via wifi, obtaining its control input to race a predefined track (see Figure \ref{fig:fulltrack}).

\subsection{Setup Limitations}
\label{apx:setuplimitations}

\rahel{\paragraph{Numerical Issues.} 
The differentiable MPC method proposed in \cite{amos2018differentiable} does not handle hard constraints in the states. Therefore, we implement a soft constraint by adding a deterministic (non-learnable) high penalty in the MPC cost. By doing this, the solution from the differentiable MPC can be slightly different from the results of standard solvers, in our simulation experiments chosen as IPOPT \cite{wachter2006implementation}, implemented using casadi \cite{Andersson2019}. Therefore, to address this during training, we exclude samples where this mismatch exceeds an arbitrary tolerance for gradient computation.}

\rahel{\paragraph{Frenet Transformation Irregularities.}
When the vehicle state has a significant lateral deviation from the centerline, it becomes increasingly susceptible to singularities, as the Frenet transformation’s stability is compromised \cite{reiter2021parameterization}. To circumvent this issue, we tightened the maximum lateral deviation, limiting the car’s range to reduce the risk of approaching these unstable regions.}

\section{Supplementary Results}
In this section, we present additional results to add more evidence to the claims we state in the main paper. Therefore, we investigate the lap time correlation with prediction horizon length in \ref{apx:laptime}, followed by execution time comparisons in \ref{apx:executiontime}, and a visualization of the learning process in \ref{apx:visualizinglearningprocess}. This is concluded by additional hardware results in \ref{apx:additionalhardwareresults}.

\subsection{Lap time with manual cost MPC}
\label{apx:laptime}

Investigating the correlation of control performance with MPC horizon length,  we provide lap times, considering the same manual cost we used for the $\mathit{MPC}_{N_S}$ and $\mathit{MPC}_{N_L}$ methods in the reported tables of the main text, for an increasing horizon length. As presented in Tables \ref{tab:laptime} and \ref{tab:laptime_paj}, the lap time decreases and seems to converge to a certain value as we increase the horizon length in both the \emph{kinematic} and \emph{Pacejka} model. The results are obtained, using the same track as we used in our training process (i.e., Tables \ref{tbl:main_kin}, \ref{tbl:main_pac}, \ref{tbl:main_im_loss}).

\begin{table}[ht]
\small
\centering
\vspace{-1.5em}
\caption{Lap time (sec) comparison for various values of $N$ using the \emph{kinematic} model. Mean and standard deviation are reported over ten runs with noise introduced on the initial state.}
\begin{tabular}{c|c}
\toprule
$N$ & Lap Time \\
\midrule
5 & $9.103 \pm 0.043$\\
10 & $8.556 \pm 0.018$ \\
15 & $8.409 \pm 0.014$ \\
20 & $8.325\pm 0.015$ \\
25 & $8.286 \pm 0.018$ \\
35 & $8.283 \pm 0.031$ \\
\bottomrule
\end{tabular}
\label{tab:laptime}
\end{table}

\begin{table}[ht]
\small
\centering
\vspace{-1.8em}
\caption{Lap time (sec) comparison for various values of $N$ using the \emph{Pacejka} model. Mean and standard deviation are reported over ten runs with noise introduced on the initial state. Results marked with $-$ did not complete the lap.}
\begin{tabular}{c|cc}
\toprule
$N$ & Lap Time \\
\midrule
6 & $-$ \\
12 & $20.361 \pm 0.201$ \\
18 & $14.088 \pm 0.132$ \\
25 & $12.171 \pm 0.133$ \\
30 & $11.634 \pm 0.187$ \\
35 & $11.088 \pm 0.070$ \\
40 & $10.350 \pm 0.212$ \\
45 & $9.894 \pm 0.100$ \\
50 & $9.615 \pm 0.071$ \\
55 & $9.477 \pm 0.036$ \\
60 & $9.417 \pm 0.016$ \\
\bottomrule
\end{tabular}
\label{tab:laptime_paj}
\end{table}

\subsection{MPC Execution time}
\label{apx:executiontime}
To illustrate the computational performance of our proposed method, we investigate its execution time employing IPOPT \cite{wachter2006implementation} for various combinations of $N_S$ and $N_L$, and both models considered. In the provided tables (Table \ref{tbl:exec_kin} and \ref{tbl:exec_pac}) it can be seen, that the inference time of our method is highly comparable to the corresponding short-horizon MPC ($\mathit{MPC}_{N_S}$). This is mainly due to the fast inference of neural networks. 

\begin{table}[ht]
\small
\centering
\vspace{-1.0em}
\caption{Execution time (ms) average for the optimization and inference process of each time step. Comparison for some values of $N_S$ and $N_L$ using the \emph{kinematic} model, considering 1000 initial states in each run (as in the imitation loss experiment). Mean and standard deviation are reported over ten runs with noise introduced on the initial state.}
\begin{tabular}{cc|ccc}
\toprule
$N_S$ & $N_L$ & $\mathit{MPC}_{N_S}$ & $\mathit{MPC}_{N_L}$ & ZipMPC  \\
\midrule
5 & 18 & $60\pm 16$ & $196\pm 51$ & $61\pm 16$ \\
5 & 25 & $60\pm 16$ & $287\pm 102$ & $61\pm 16$ \\
10 & 18 & $102\pm 32$ & $196\pm 51$ & $103\pm 32$ \\
10 & 25 & $102\pm 32$ & $287\pm 102$ & $103\pm 32$ \\
\bottomrule
\label{tbl:exec_kin}
\end{tabular}
\end{table}

\begin{table}[ht]
\small
\centering
\vspace{-2.2em}
\caption{Execution time (ms) average for the optimization and inference process of each time step. Comparison for some values of $N_S$ and $N_L$ using the \emph{Pacejka} model, considering 1000 initial states in each run (as in the imitation loss experiment). Mean and standard deviation are reported over ten runs with noise introduced on the initial state.}
\begin{tabular}{cc|ccc}
\toprule
$N_S$ & $N_L$ & $\mathit{MPC}_{N_S}$ & $\mathit{MPC}_{N_L}$ & ZipMPC  \\
\midrule
6 & 35 & $65\pm 59$ & $476\pm 211$ & $66\pm 59$ \\
6 & 45 & $65\pm 59$ & $704\pm 337$ & $66\pm 59$ \\
12 & 35 & $132\pm 56$ & $476\pm 211$ & $133\pm 56$ \\
12 & 45 & $132\pm 56$ & $704\pm 337$ & $133\pm 56$ \\
\bottomrule
\label{tbl:exec_pac}
\end{tabular}
\vspace{-2.2em}
\end{table}

\subsection{Visualizing the learning process}
\label{apx:visualizinglearningprocess}
\begin{figure*}[h]
\vspace{-1.0em}
    \centering    \includegraphics[width=\linewidth]{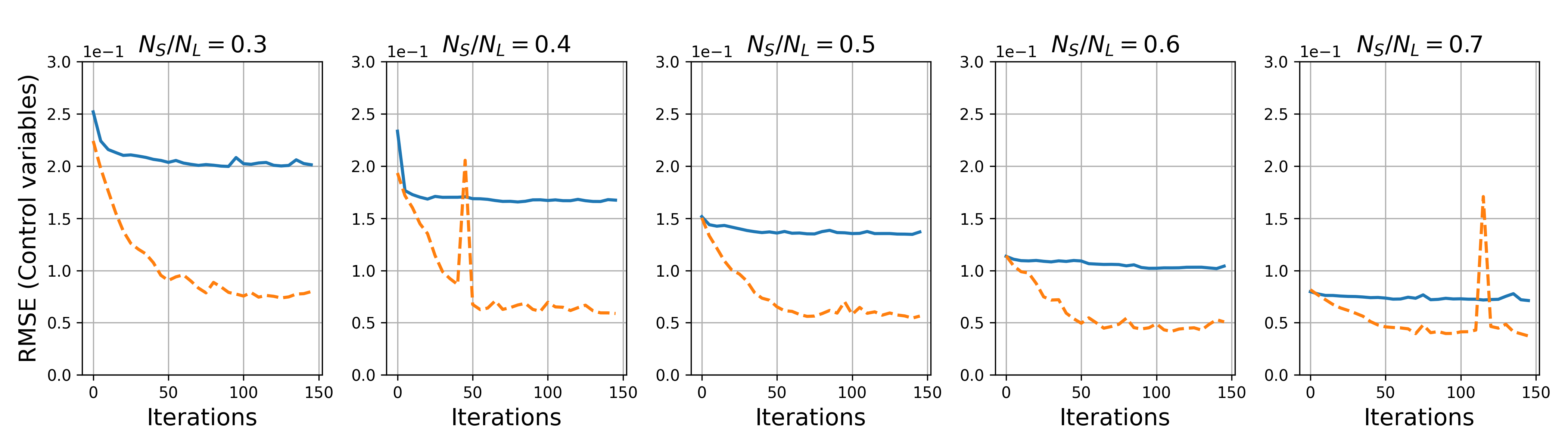}
    \vspace{-1.5em}
    \caption{Validation curve comparison of context-free and context-aware ZipMPC across different ratios of short (learned) and long-horizons ($N_S/N_L$). This graph is analogous to Figure \ref{fig:context_on_rmse} but shows the full learning curve within the validation set. $N_L=20$ was considered in this experiment for the \emph{kinematic} model.}
    \vspace{-1.5em}
\label{fig:context_curves}
\end{figure*}

\begin{figure*}[h]
    \centering    \includegraphics[width=\linewidth]{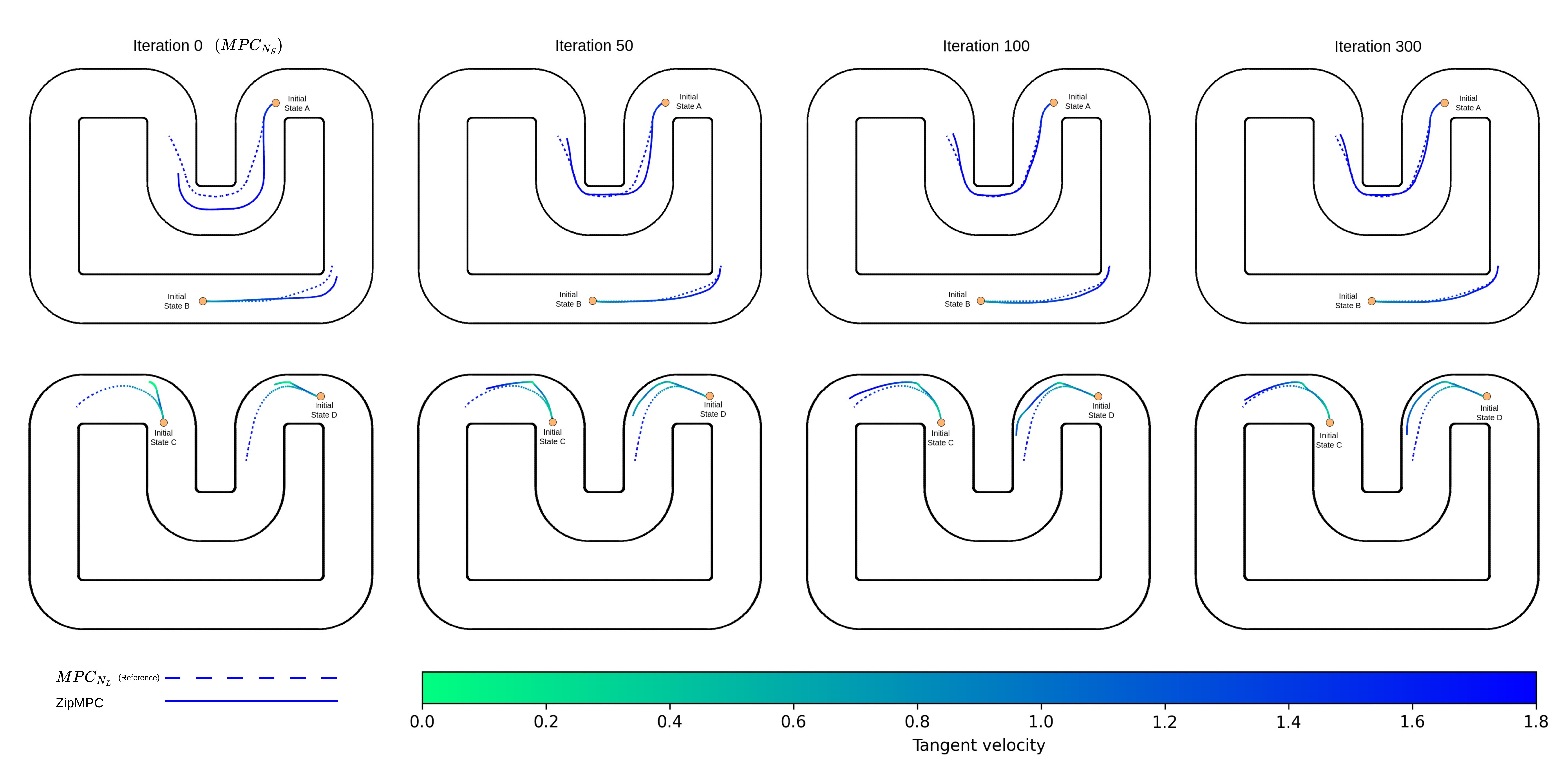}
    \vspace{-1.5em}
    \caption{Comparison of trajectories inferred by our framework and the target trajectory from $\mathit{MPC}_{N_L}$. Top and bottow row illustrate two trajectories with different initial states using the \emph{kinematics bycicle model} and the \emph{Pacejka bycicle model}, respectively. Each column depicts the trajectory after training the model for 0, 50, 100, and 300 iterations.}
    \label{fig:it_samples}
    \vspace{-1.5em}
\end{figure*}

We analyze the effect of varying initial states in the \textit{Pacejka} model to evaluate the robustness of our framework under diverse initial states and to compare its performance against the $\mathit{MPC}_{N_S}$ and $\mathit{MPC}_{N_L}$ baselines. Specifically, we initialize our model in various initial state configurations and observe its ability to approximate the "target" trajectory produced by $\mathit{MPC}_{N_L}$. To illustrate the progression of learning, we plot results at different stages of training: at 0 iterations (approximately mimicking the $\mathit{MPC}_{N_S}$ method), and after 20, 50, and 100 iterations. The results provided in Figure \ref{fig:it_samples} demonstrate that as the number of iterations increases, the trajectories generated by our model progressively align more closely with the target trajectory, highlighting the effectiveness of the learning process in refining the trajectory over time. Furthermore, in Figure \ref{fig:context_curves}, we augment the results illustrated in Figure \ref{fig:context_on_rmse}, visualizing the full learning curves of context-free and context-aware ZipMPC across different ratios of short (learned) and long-horizons ($N_S/N_L$). It is evident from this that the incorporation of context information enhances the learning performance in all of the selected scenarios. 

\subsection{Additional hardware results}
\label{apx:additionalhardwareresults}
\begin{figure}[h!]
\vspace{-0.0em}
\centering
\begin{subfigure}{0.19\textwidth}
    \includegraphics[width=\textwidth,trim={5cm 0 5cm 0},clip]{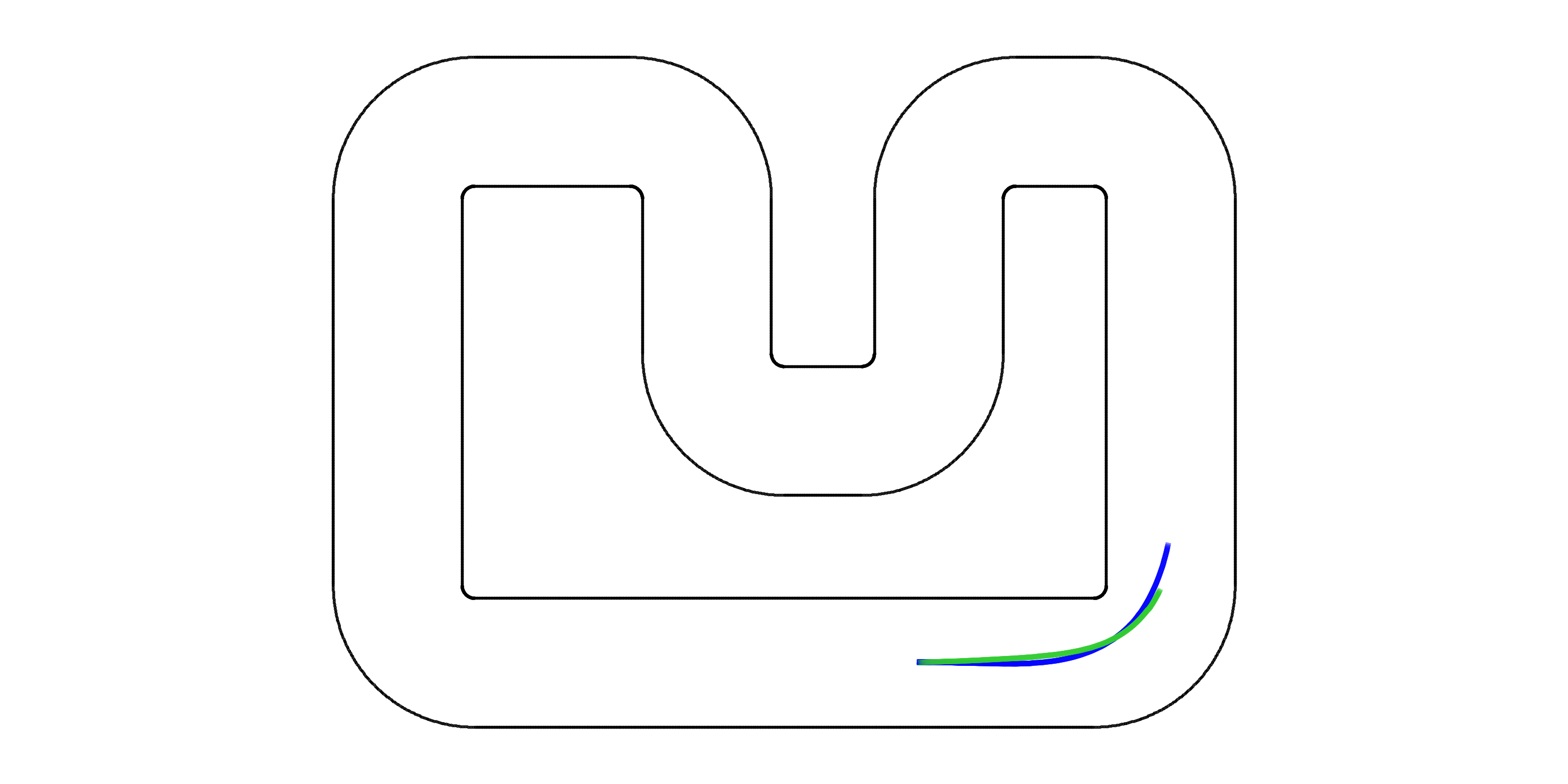}
    \caption{1sec.}
    \label{fig:first}
\end{subfigure}
\hfill
\begin{subfigure}{0.19\textwidth}
    \includegraphics[width=\textwidth,trim={5cm 0 5cm 0},clip]{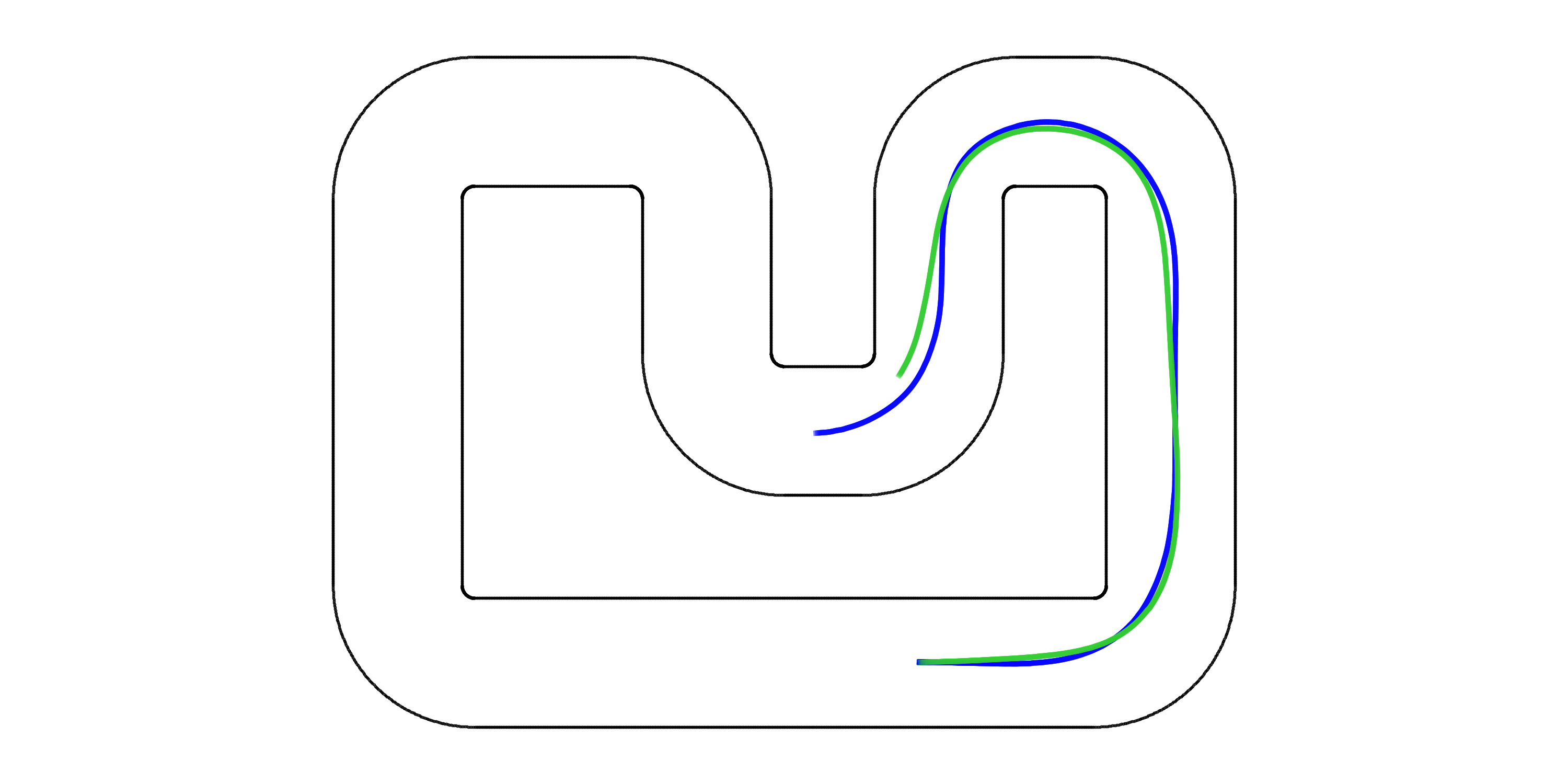}
    \caption{3sec.}
    \label{fig:second}
\end{subfigure}
\hfill
\begin{subfigure}{0.19\textwidth}
    \includegraphics[width=\textwidth,trim={5cm 0 5cm 0},clip]{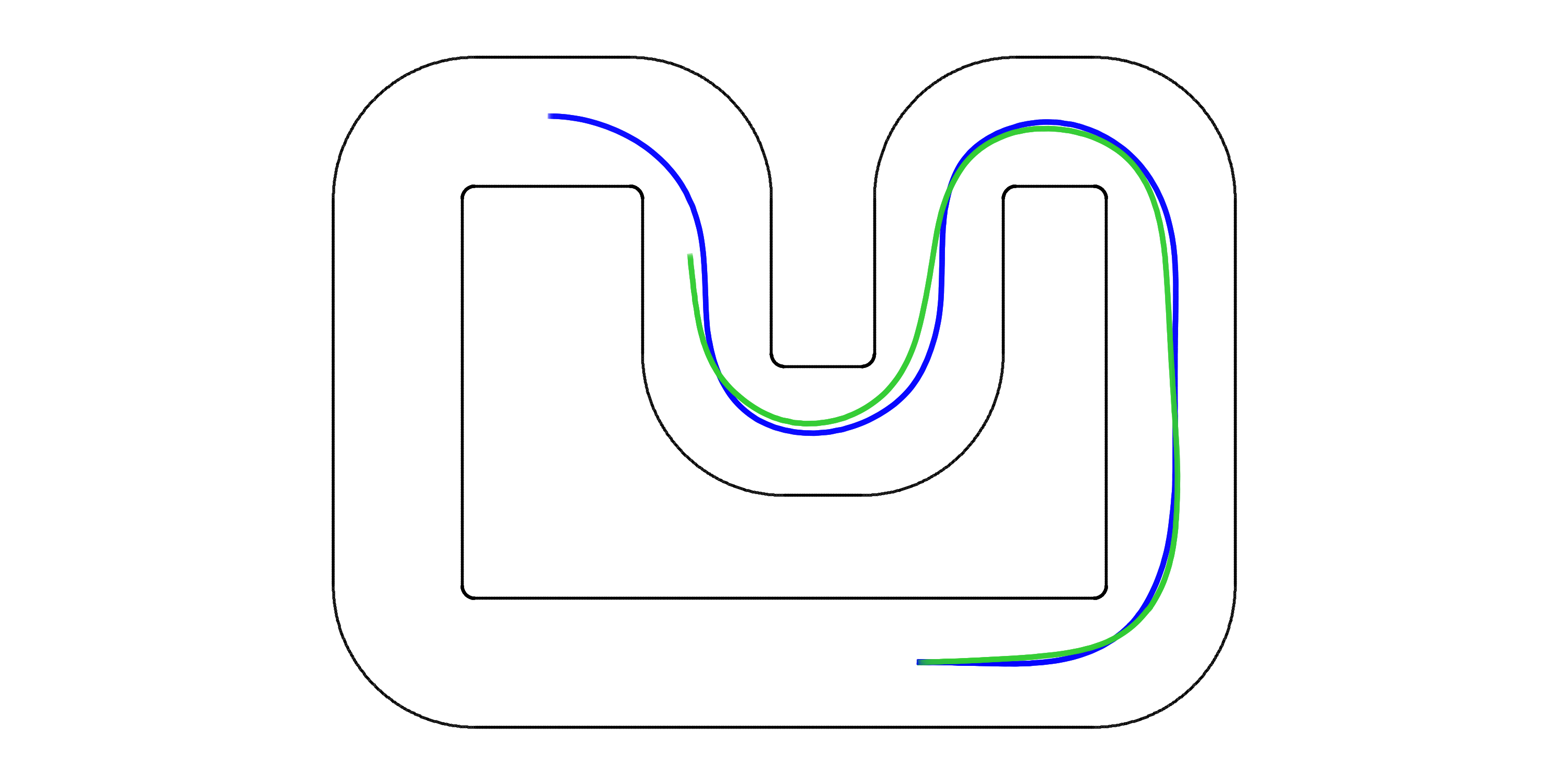}
    \caption{4sec.}
    \label{fig:second}
\end{subfigure}
\hfill
\begin{subfigure}{0.19\textwidth}
    \includegraphics[width=\textwidth,trim={5cm 0 5cm 0},clip]{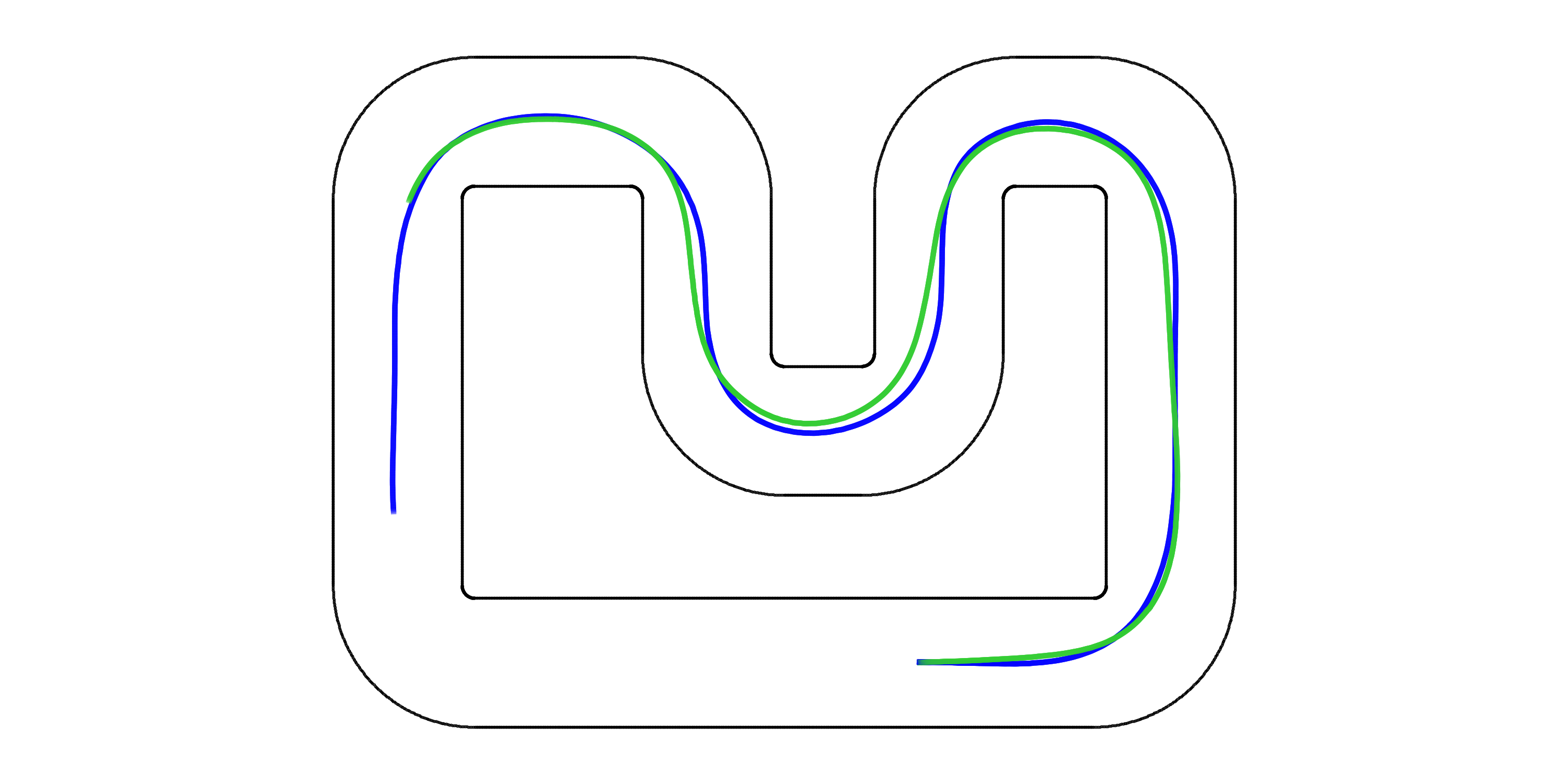}
    \caption{5sec.}
    \label{fig:third}
\end{subfigure}
\hfill
\begin{subfigure}{0.19\textwidth}
    \includegraphics[width=\textwidth,trim={5cm 0 5cm 0},clip]{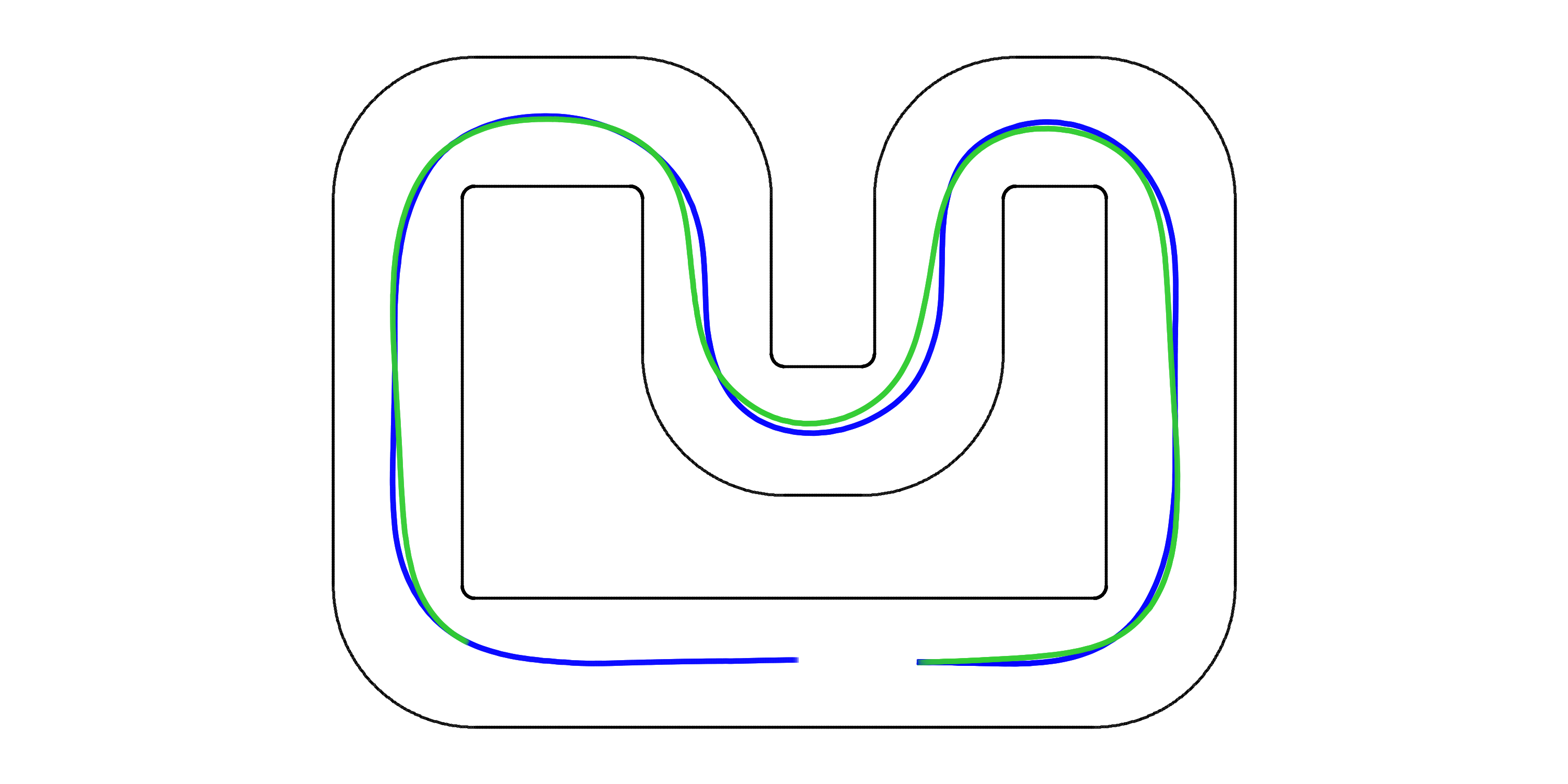}
    \caption{6sec.}
    \label{fig:third}
\end{subfigure}

\caption{Comparison of ZipMPC (green) and $\mathit{MPC}_{N_L}$ (blue) for different instances in time, executed on hardware. Thereby $N_S$ is chosen as 20 and $N_L$ is set equal to 40.}
\label{fig:hardware_20_40_long_comp}
\vspace{-1em}
\end{figure}

\begin{figure}[h!]
\vspace{-0.0em}
\centering
\begin{subfigure}{0.19\textwidth}
    \includegraphics[width=\textwidth,trim={5cm 0 5cm 0},clip]{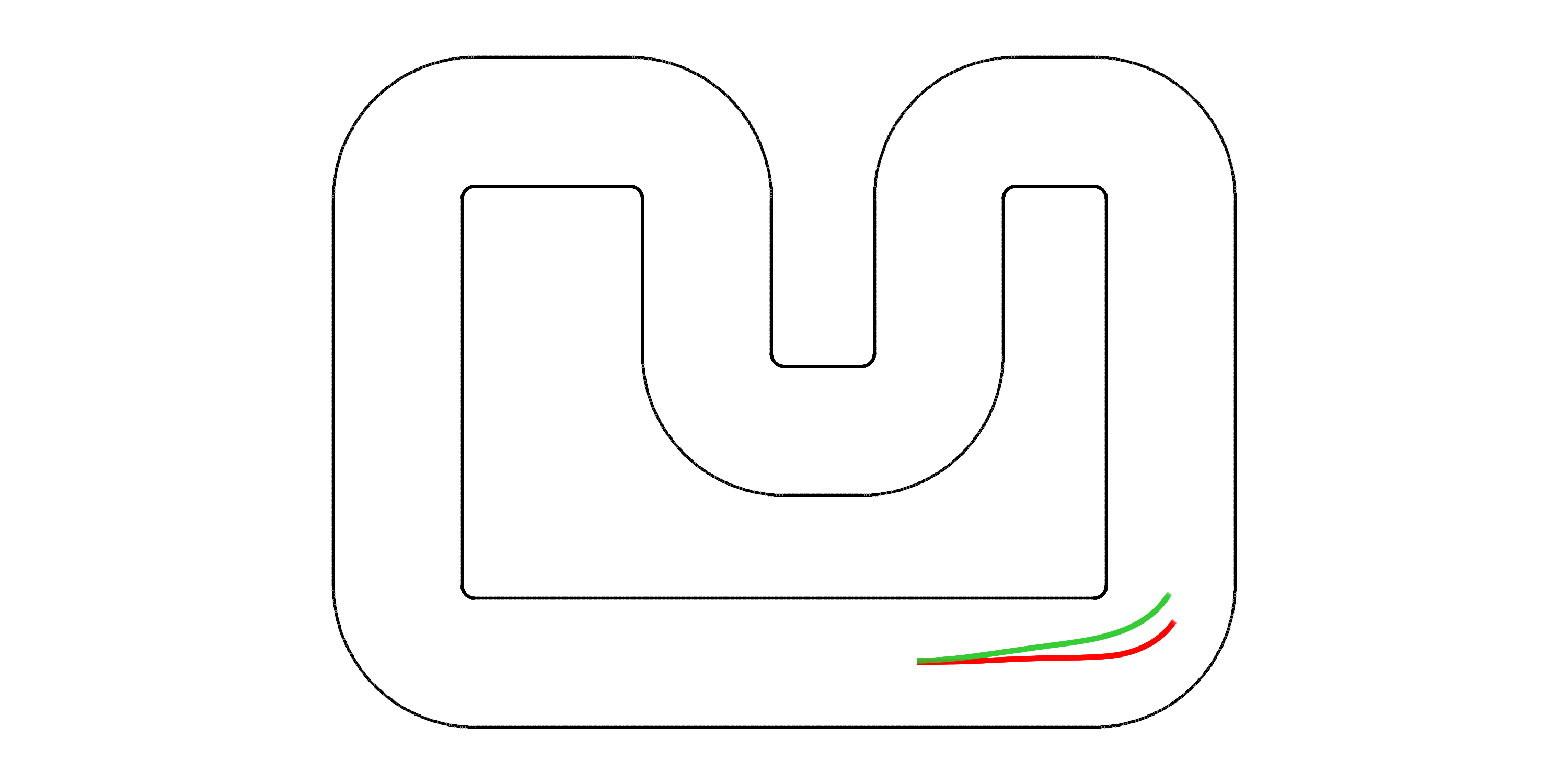}
    \caption{1sec.}
    \label{fig:first}
\end{subfigure}
\hfill
\begin{subfigure}{0.19\textwidth}
    \includegraphics[width=\textwidth,trim={5cm 0 5cm 0},clip]{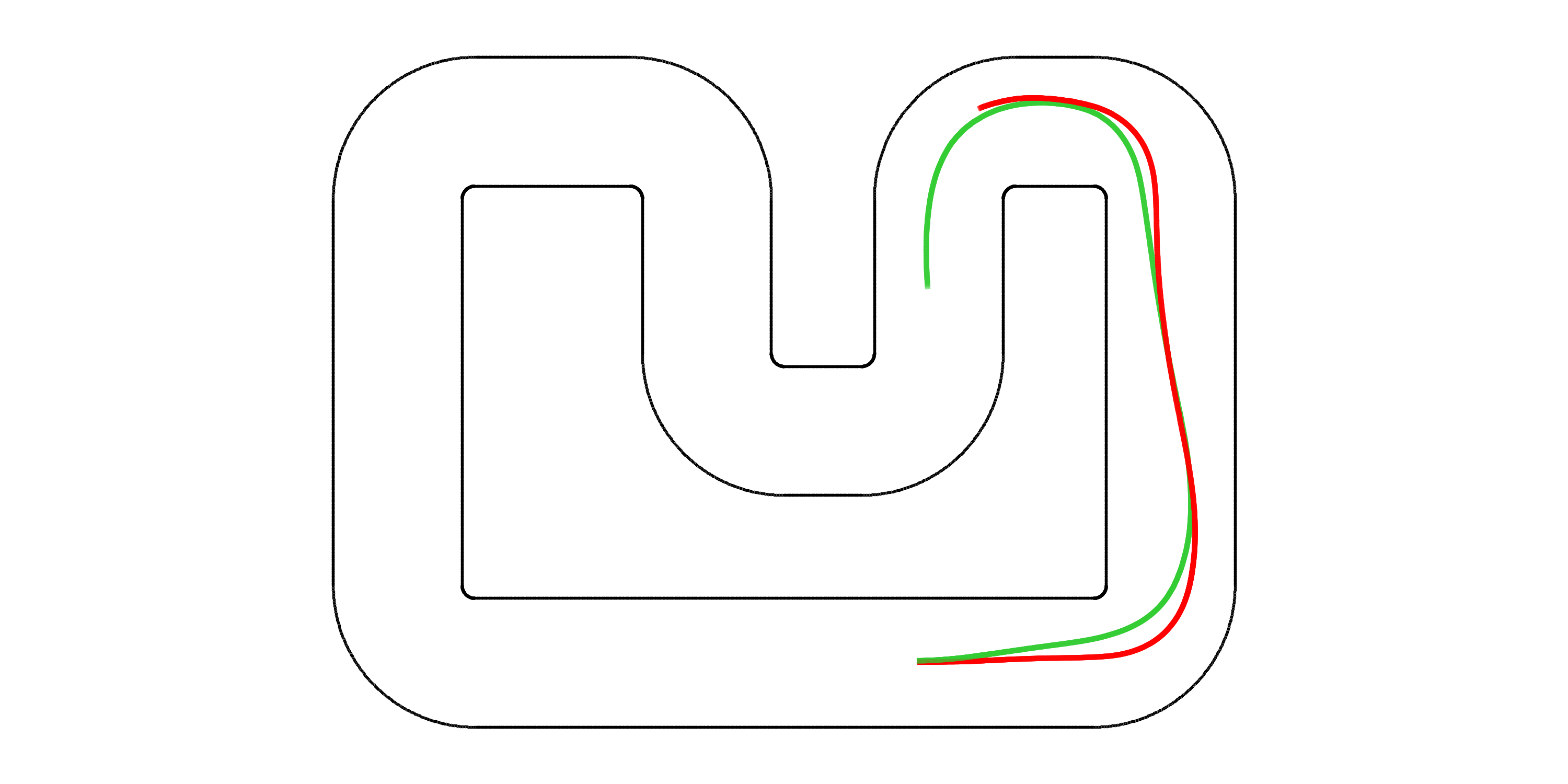}
    \caption{3sec.}
    \label{fig:second}
\end{subfigure}
\hfill
\begin{subfigure}{0.19\textwidth}
    \includegraphics[width=\textwidth,trim={5cm 0 5cm 0},clip]{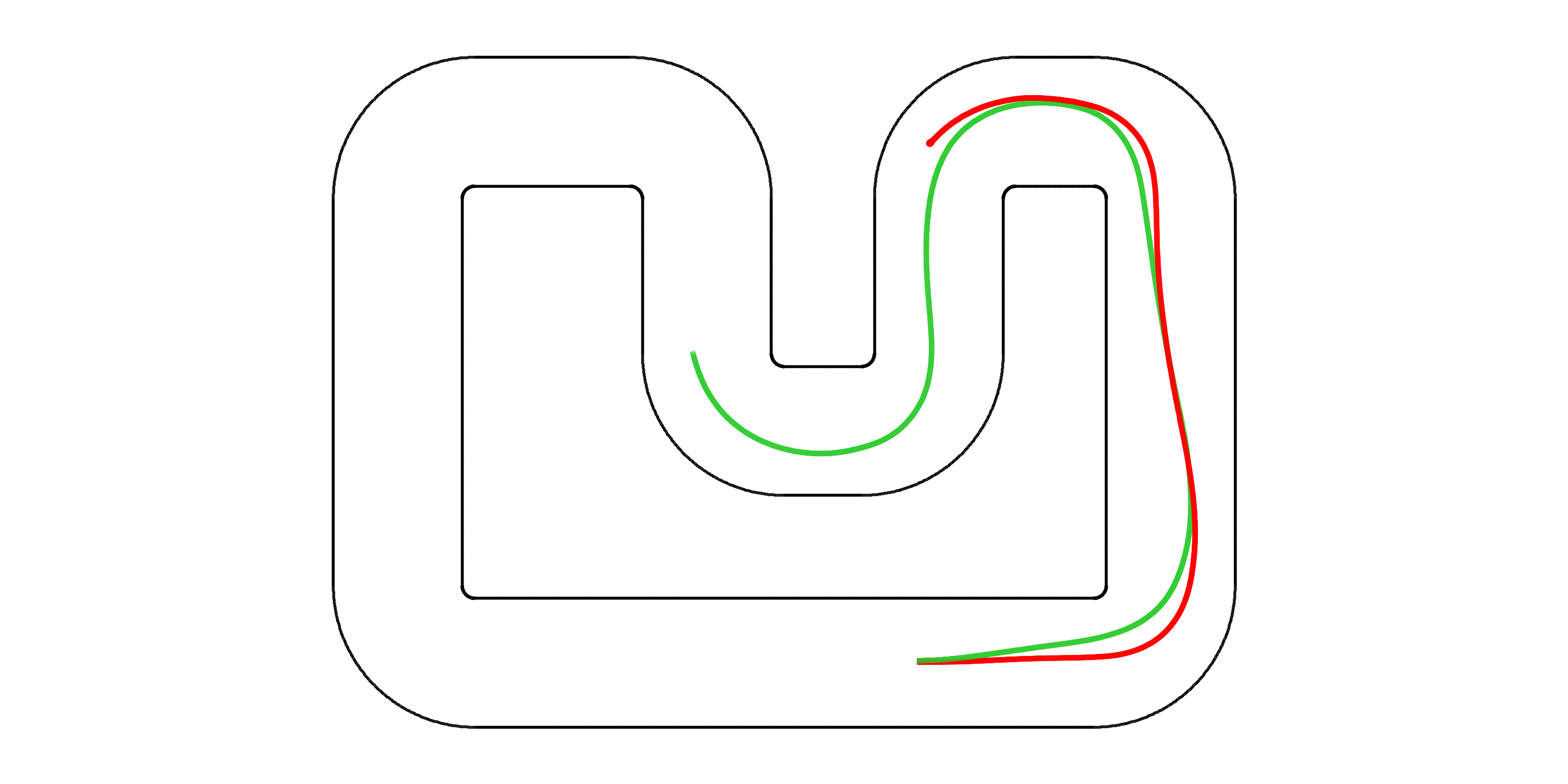}
    \caption{4sec.}
    \label{fig:second}
\end{subfigure}
\hfill
\begin{subfigure}{0.19\textwidth}
    \includegraphics[width=\textwidth,trim={5cm 0 5cm 0},clip]{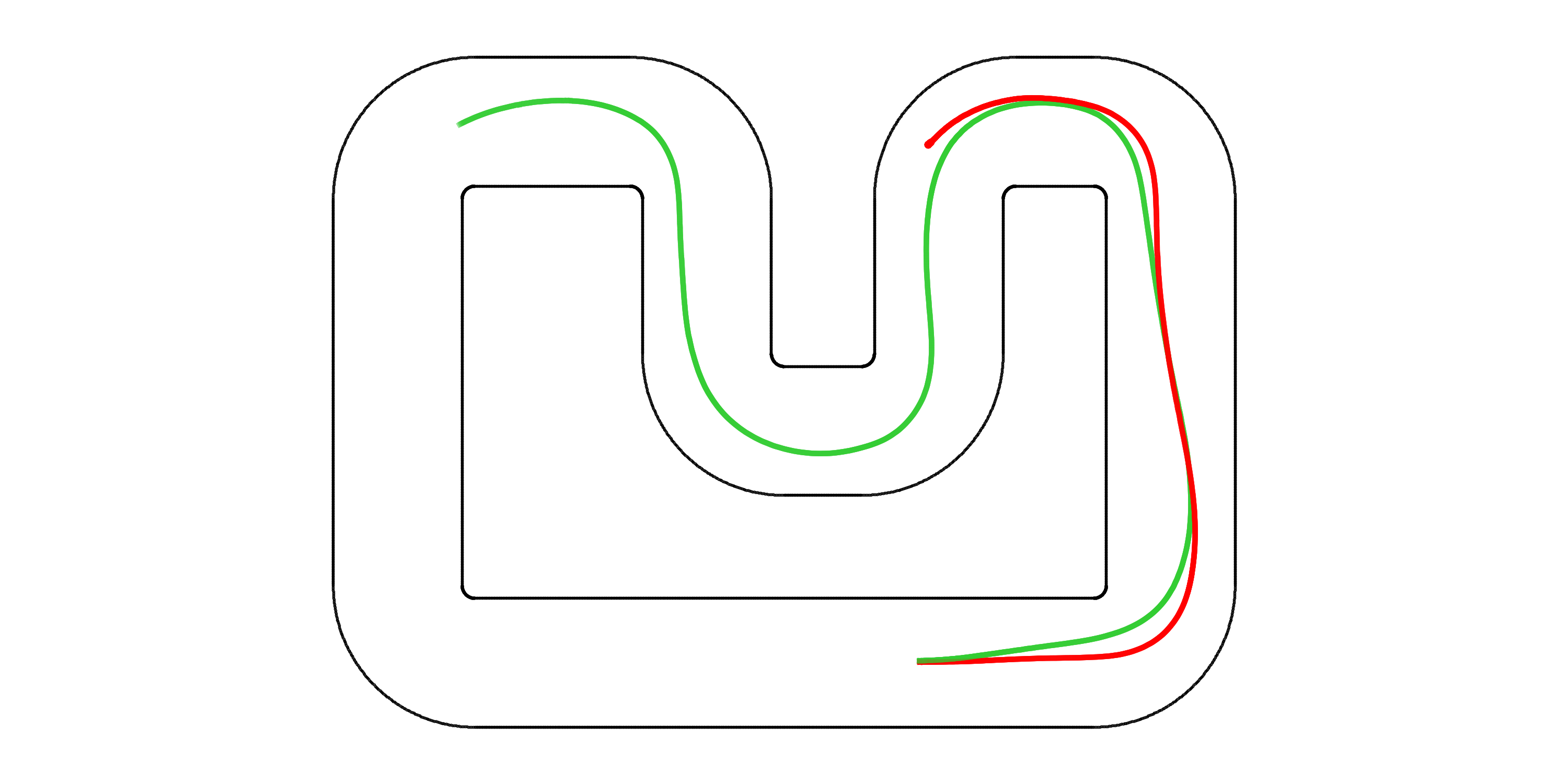}
    \caption{5sec.}
    \label{fig:third}
\end{subfigure}
\hfill
\begin{subfigure}{0.19\textwidth}
    \includegraphics[width=\textwidth,trim={5cm 0 5cm 0},clip]{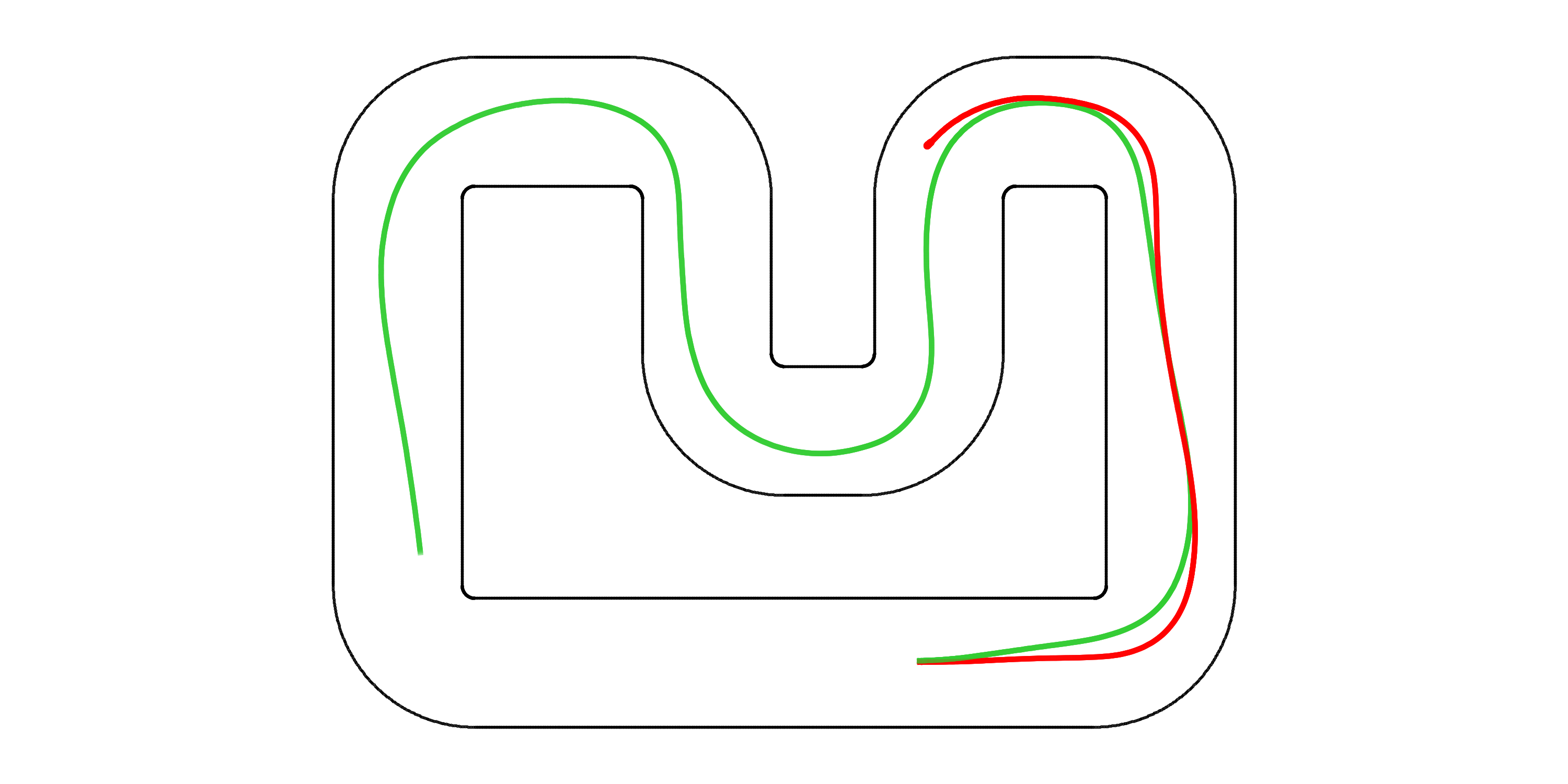}
    \caption{6sec.}
    \label{fig:third}
\end{subfigure}
        
\caption{Comparison of ZipMPC (green) and $\mathit{MPC}_{N_S}$ (red) for different instances in time, executed on hardware. Thereby $N_S$ is chosen as 10 and $N_L$ is set equal to 25.}
\label{fig:hardware_10_25}
\vspace{-1em}
\end{figure}

\begin{figure}[h!]
\vspace{-0.0em}
\centering
\begin{subfigure}{0.19\textwidth}
    \includegraphics[width=\textwidth,trim={5cm 0 5cm 0},clip]{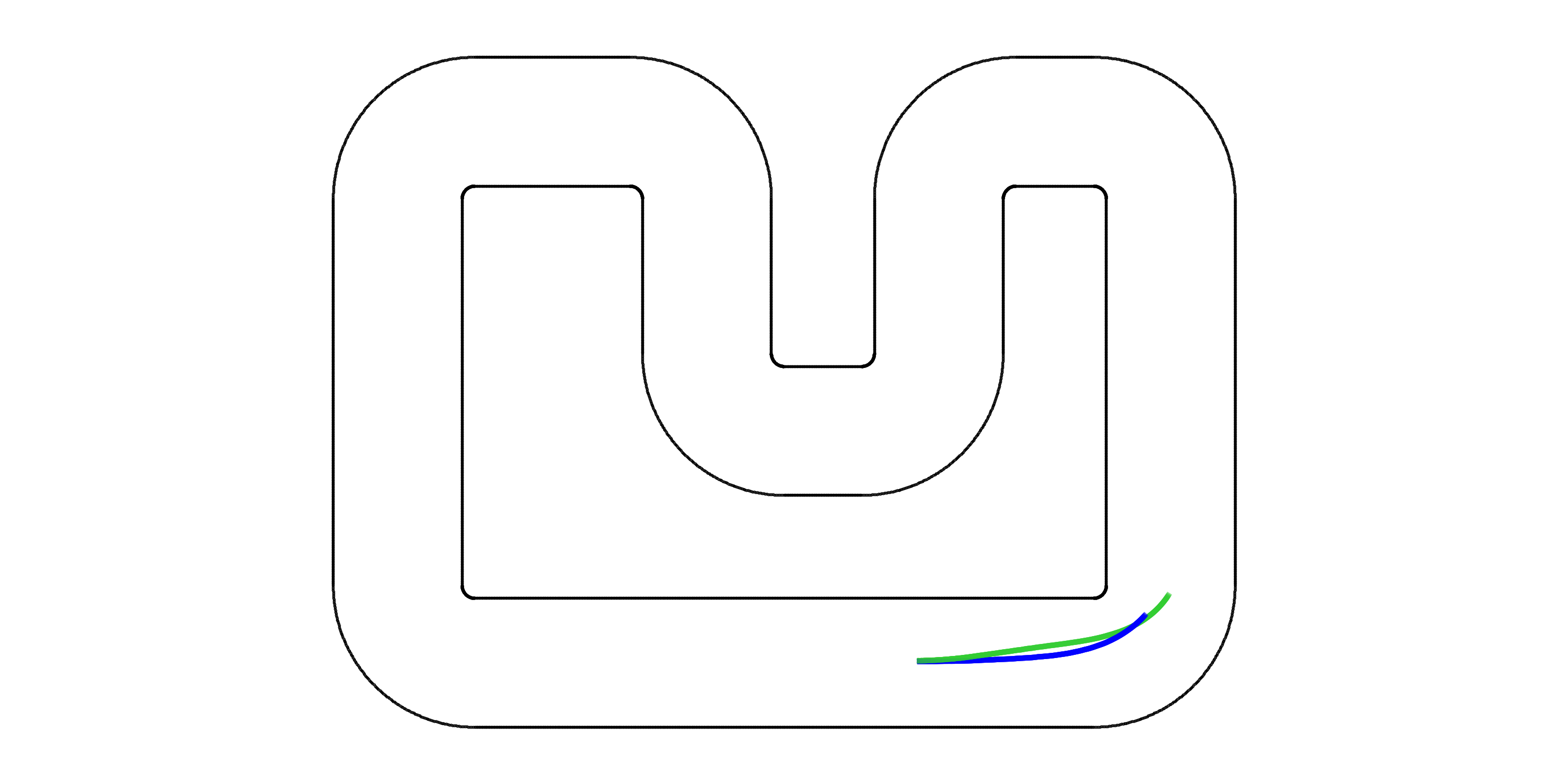}
    \caption{1sec.}
    \label{fig:first}
\end{subfigure}
\hfill
\begin{subfigure}{0.19\textwidth}
    \includegraphics[width=\textwidth,trim={5cm 0 5cm 0},clip]{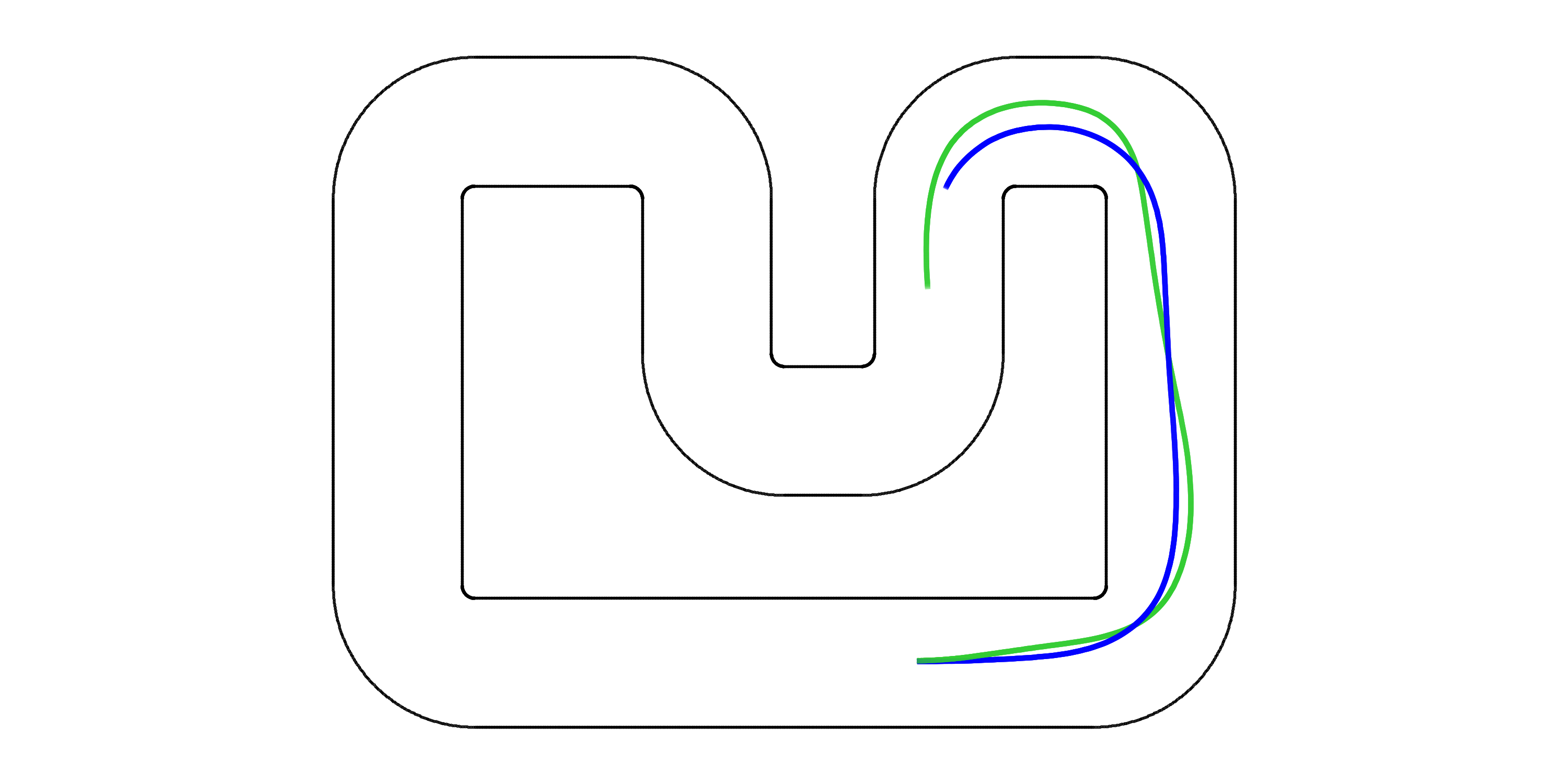}
    \caption{3sec.}
    \label{fig:second}
\end{subfigure}
\hfill
\begin{subfigure}{0.19\textwidth}
    \includegraphics[width=\textwidth,trim={5cm 0 5cm 0},clip]{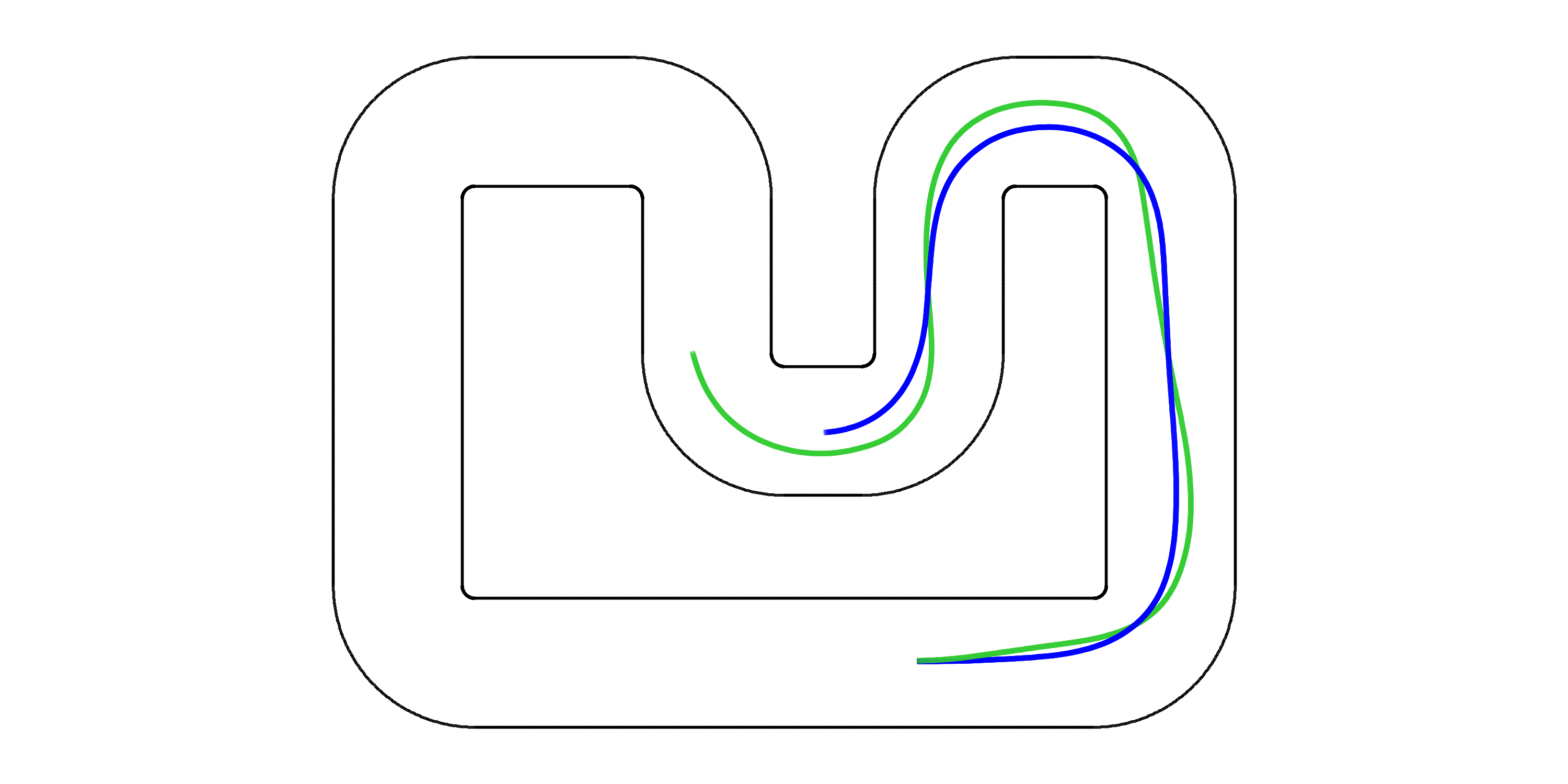}
    \caption{4sec.}
    \label{fig:second}
\end{subfigure}
\hfill
\begin{subfigure}{0.19\textwidth}
    \includegraphics[width=\textwidth,trim={5cm 0 5cm 0},clip]{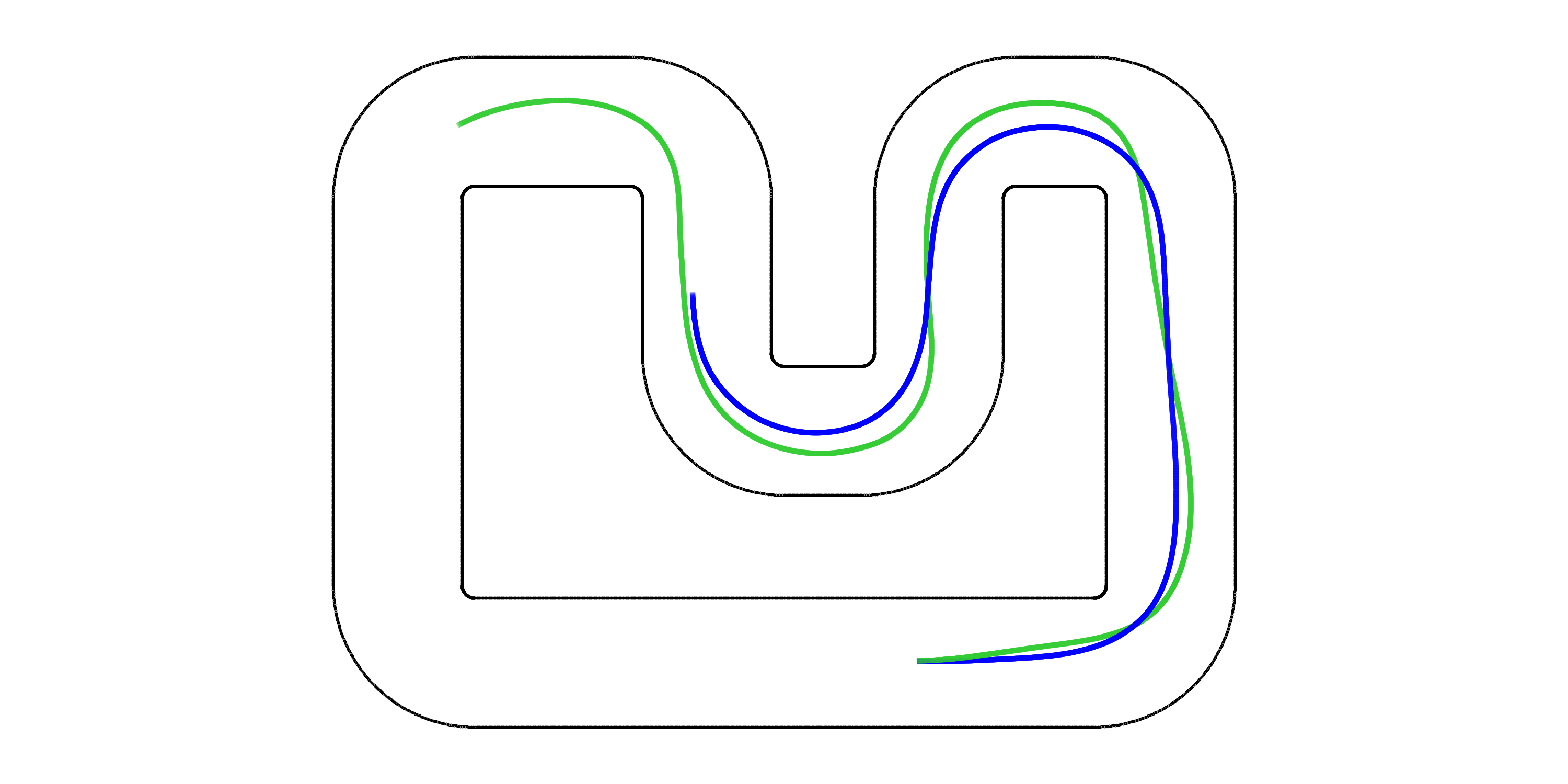}
    \caption{5sec.}
    \label{fig:third}
\end{subfigure}
\hfill
\begin{subfigure}{0.19\textwidth}
    \includegraphics[width=\textwidth,trim={5cm 0 5cm 0},clip]{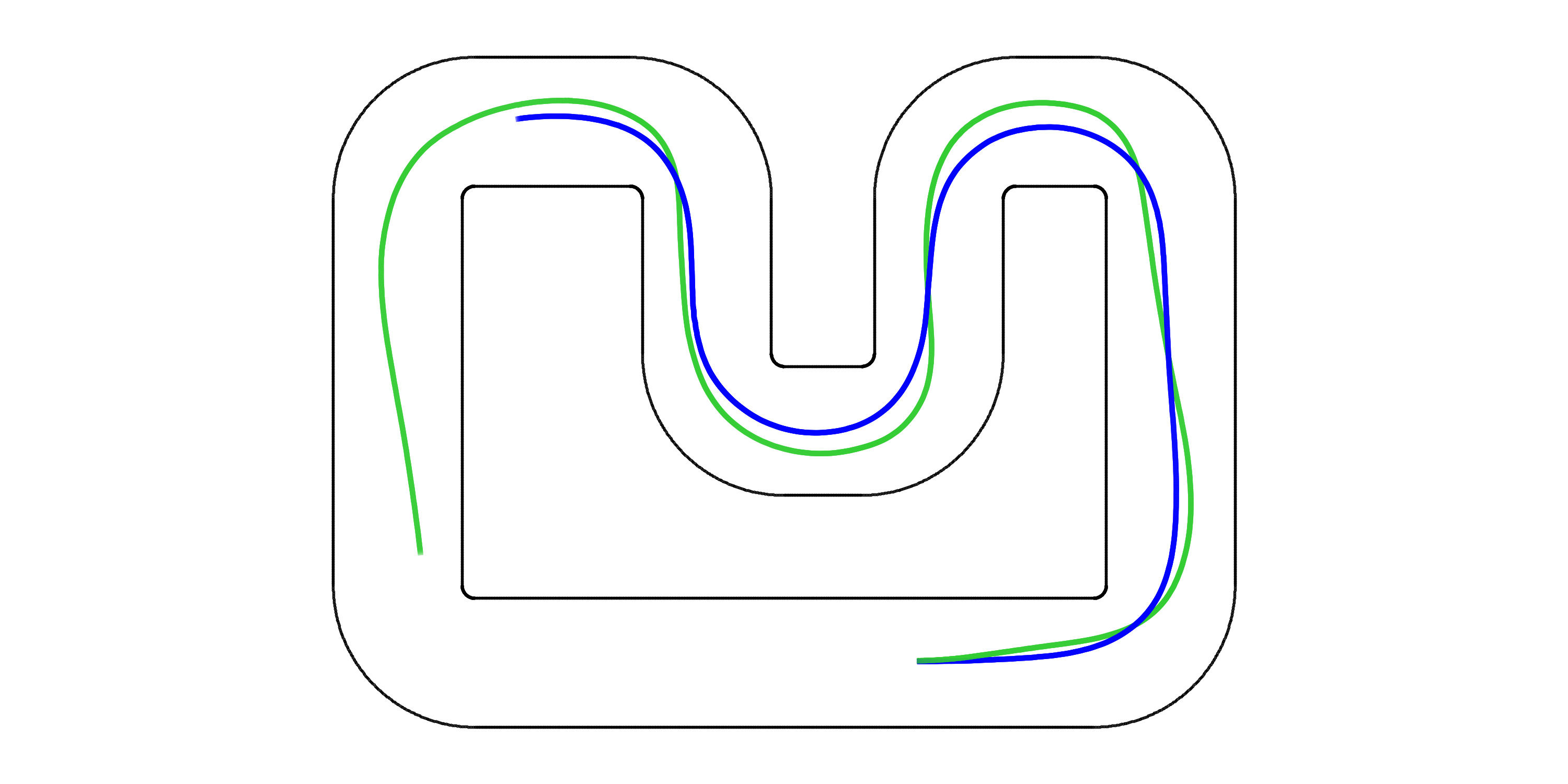}
    \caption{6sec.}
    \label{fig:third}
\end{subfigure}
        
\caption{Comparison of ZipMPC (green) and $\mathit{MPC}_{N_L}$ (blue) for different instances in time, executed on hardware. Thereby $N_S$ is chosen as 10 and $N_L$ is set equal to 25.}
\label{fig:hardware_10_25_long_comp}
\vspace{-1em}
\end{figure}

For the presented horizon length combination in Section \ref{sec:hardware}, we further illustrate a comparison of the control behaviour of our learned ZipMPC and $\mathit{MPC}_{N_L}$ for the same instances in time. The results are visualized in Figure \ref{fig:hardware_20_40_long_comp}. While the presented choice of $N_S = 20$ and $N_L = 40$ allows for a successful lap completion of both, our proposed ZipMPC method and $\mathit{MPC}_{N_S}$, we further tested our proposed method with values of $N_S$ for which $\mathit{MPC}_{N_S}$ is not capable of finishing the lap, i.e., we tested the horizon length combination $N_S = 10$ and $N_L = 25$. The comparison of our proposed ZipMPC method with $\mathit{MPC}_{N_S}$ is illustrated in Figure \ref{fig:hardware_10_25} and Figure \ref{fig:hardware_10_25_long_comp} visualizes the comparison of ZipMPC with the reference $\mathit{MPC}_{N_L}$. It can be seen that while the application with $\mathit{MPC}_{N_S}$ gets stuck after 4 seconds in the third corner, the corresponding ZipMPC implementation not only outperforms the short-horizon MPC in the first few seconds but also successfully completes the lap. Furthermore, do the slight inaccuracies of the learned imitation behaviour, as well as the faster computation time, lead to an outperformance of $\mathit{MPC}_{N_L}$.

\end{document}